\documentclass[acmsmall]{acmart}
\setcopyright{acmlicensed}
\acmJournal{TIST}
\acmYear{2022}
\acmVolume{1}
\acmNumber{1}
\acmArticle{1}
\acmMonth{1}
\acmPrice{15.00}
\acmDOI{10.1145/3511711}

\usepackage{balance}
\usepackage{amsmath,amsfonts}
\usepackage{algorithmic}
\usepackage{graphicx}
\usepackage{textcomp}
\usepackage{xcolor}
\usepackage{xcolor}
\usepackage{xspace}
\usepackage[ruled, vlined, linesnumbered]{algorithm2e}
\usepackage{algorithmic}
\usepackage{graphicx}
\usepackage{amsmath}
\usepackage{amsfonts}
\usepackage{subcaption}
\usepackage{amsthm}
\usepackage{mathrsfs}
\usepackage{booktabs}
\usepackage{color}
\usepackage{hyperref}
\usepackage{lipsum}

\newcommand{\ourm}{\textsc{Axolotl}\xspace}
\newcommand{\ours}{\textsc{Axo}\xspace}

\theoremstyle{definition}
\newtheorem*{problem*}{Problem Statement}%[section]

\newcommand{\bm}[1]{\mathbf{#1}}
\newcommand{\cm}[1]{\mathcal{#1}}
\newcommand{\ds}[1]{\boldsymbol{\mathcal{#1}}}
\newcommand{\bs}[1]{\boldsymbol{#1}}
\newcommand{\norm}[1]{\Big \lVert #1 \Big \rVert}
\newcommand{\snorm}[1]{\big \lVert #1 \big \rVert}

\usepackage{colortbl}

\renewcommand{\comment}[1]{}
\usepackage{xcolor}
\usepackage{multirow,amsmath,booktabs,dcolumn}
\usepackage{todonotes}
\usepackage{colortbl}
\usepackage{paralist}
\usepackage{graphicx}
\usepackage{caption}

\usepackage{mathtools}

\usepackage{xcolor,colortbl}
\newcolumntype{a}{>{\columncolor{blue!5}}c}
\newcolumntype{b}{>{\columncolor{red!5}}c}

\begin{document}
\title{Doing More with Less: Overcoming Data Scarcity for POI Recommendation via Cross-Region Transfer}
\author{Vinayak Gupta}
\affiliation{
  \institution{Indian Institute of Technology Delhi}
  \city{New Delhi}
  \country{India}}
  \email{vinayak.gupta@cse.iitd.ac.in}

\author{Srikanta Bedathur}
\affiliation{
  \institution{Indian Institute of Technology Delhi}
  \city{New Delhi}
  \country{India}}
  \email{srikanta@cse.iitd.ac.in}
\renewcommand{\shorttitle}{POI Recommendation via Cross-Region Transfer}

\begin{abstract}
Variability in social app usage across regions results in a high skew of the quantity and the quality of check-in data collected, which in turn is a challenge for effective location recommender systems. In this paper, we present \ourm (Automated \textit{cross} Location-network Transfer Learning), a novel method aimed at transferring location preference models learned in a data-rich region to significantly boost the quality of recommendations in a data-scarce region.
%
%
%As the user base for social apps varies with different regions, so does the volume of trajectory data. Innately, there are regions with high and simultaneously low-quality check-in data. In this paper, we present \ourm (Automated \textit{cross} Location-network Transfer Learning) that for the first time determines, how to utilize the location preference knowledge from a data-rich region for recommendations in a data-scarce region. 
%
%
%Despite the widespread use of services such as Foursquare for capturing mobility data, recent reports suggest a large fraction of data collected is either unreliable or unusable due to privacy rights of users. 
%This loss of data poses a serious challenge to the existing approaches that model the user mobility preferences since they critically rely on large volumes of high quality data and thus perform poorly in regions where the data is particularly sparse. In this paper, we address these issues and present a novel framework called \ourm, which shows superior location recommendation performance even in regions with very limited data. 
\ourm predominantly deploys two channels for information transfer, (1) a \emph{meta-learning} based procedure learned using location recommendation as well as social predictions, and (2) a \textit{lightweight} unsupervised cluster-based transfer across users and locations with similar preferences. Both of these work together synergistically to achieve improved accuracy of recommendations in data-scarce regions without any prerequisite of overlapping users and with minimal fine-tuning. We build \ourm on top of a \emph{twin graph-attention} neural network model used for capturing the user- and location-conditioned influences in a user-mobility graph for each region. %For a specific region, we capture the disparate user- and location-conditioned influences via a user mobility graph with a \emph{twin graph-attention} model.
We conduct extensive experiments on 12 user mobility datasets across the U.S., Japan, and Germany, using 3 as \textit{source} regions and 9 of them (that have much sparsely recorded mobility data) as \textit{target} regions. Empirically, we show that \ourm achieves up to 18\% better recommendation performance than the existing state-of-the-art methods across all metrics.
\end{abstract}



\maketitle

\section{Introduction}\label{intro}
\begin{figure}[t!]
\begin{subfigure}[]{0.30\columnwidth}
\centering
\includegraphics[width=\linewidth]{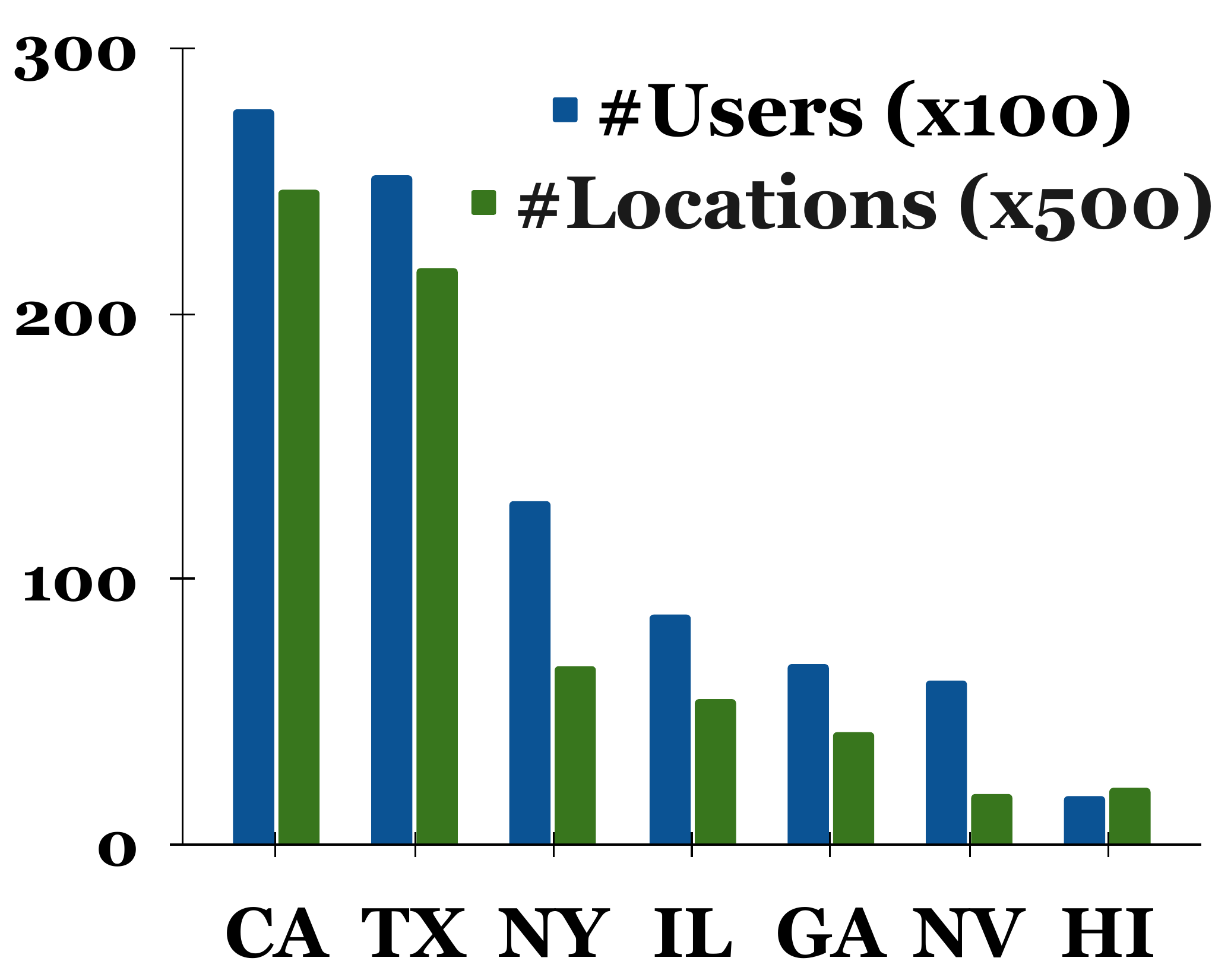}
\caption{State-wise Data}
\label{fig:1a}
\end{subfigure}\hfill
\begin{subfigure}[]{0.30\columnwidth}
\centering
\includegraphics[width=\linewidth, height=3.2cm]{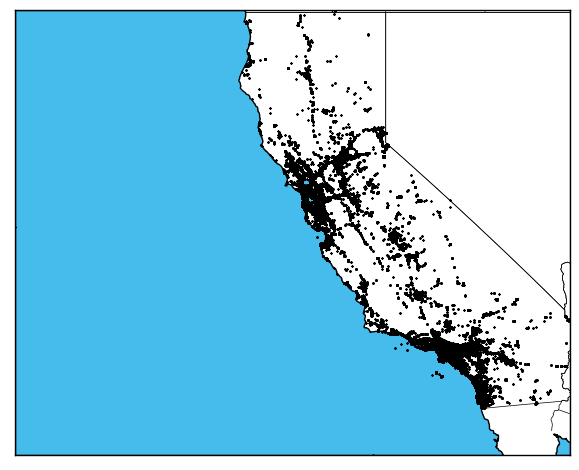}
\caption{California}
\label{fig:1b}
\end{subfigure}\hfill
\begin{subfigure}[]{0.30\columnwidth}
\centering
\includegraphics[width=\linewidth, height=3.2cm]{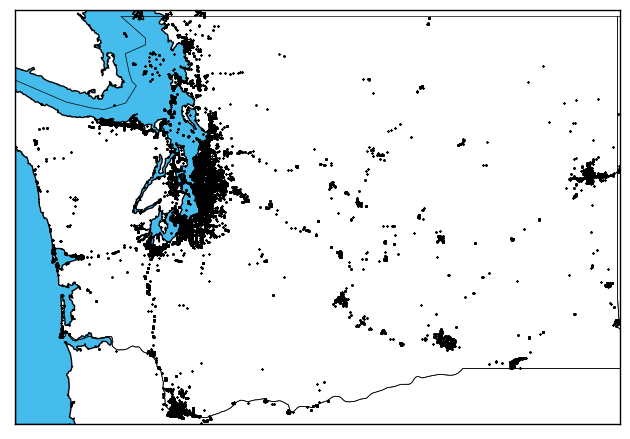}
\caption{Washington}
\label{fig:1c}
\end{subfigure}

\caption{Skew in the volume of mobility data across different states in the US (Figure~\ref{fig:1a}) and the large variation in region-specific density of mobility data between California and Washington in Figures ~\ref{fig:1b} and ~\ref{fig:1c} respectively (based on Gowalla dataset~\cite{scellato}).}
\label{fig:distri}
\end{figure}

As POI (Points-of-Interest) gathering services such as Foursquare, Yelp, and Google Places are becoming widespread, there is significant research in extracting location preferences of users to predict their mobility behavior and recommend next POIs that users are likely to visit~\cite{cara, locate, cheng2013you, deepmove, deepst, ruirui}. However, the quality of POI recommendations for users in regions where there is a severe scarcity of mobility data is much poorer in comparison with those from data-rich regions. This is a critical problem affecting the state-of-the-art approaches~\cite{regiontrans, www, similarwww}.
%%%We address the problems associated with inefficient POI recommendation for users in data-scarce regions that are severely affected by the variation in cross-region social app usage in contrast to data-rich regions. %while transferring trajectory knowledge from data-rich source regions without any presumption of inter-region user overlap. 
%%%This is a paramount problem with POI recommendation systems as the current state-of-the-art approaches owe their performance to the widespread availability of high-quality mobility data \cite{cara, locate, cheng2013you, deepmove, deepst, ruirui}.
%Understanding the complex nature of these datasets is critical for point-of-interest (POI) recommendation~\cite{cara, locate}, next check-in prediction~\cite{cheng2013you, deepmove}, locating prospective customers \cite{advlbsn, ruirui} and applications dependent on location-awareness~\cite{tinder, deepst, lichman}.
The situation is further exacerbated in the recent times due to the advent of various restrictions for collecting personal data and growing awareness (in some geopolitical regions) about the need for personal privacy~\cite{privacy, privacy2}. It not only means that there is overall reduction in the high quality (useful) data\footnote{Nearly 80\% of the data generated by Foursquare users is discarded~\cite{fsq}.}, but, even more importantly, it introduces a \emph{high-skew} in the mobility data across different regions (see Figure~\ref{fig:distri}) -- primarily due to varying views towards personal privacy across these regions. \\
%Advent of further buoyed by GDPR\footnote{\href{https://gdpr-info.eu/}{https://gdpr-info.eu/}} and CCPA\footnote{\href{https://oag.ca.gov/privacy/ccpa}{https://oag.ca.gov/privacy/ccpa}} laws and users voluntarily hiding their mobility \cite{privacy} can result in further data scrubbing.
%This not only means that there is less data to learn from, but it also exacerbates the \emph{high skew} that already exists in the user mobility data statistics across different regions.
%
%To highlight this issue, consider the statistics of users and their mobility information given in Figure~\ref{fig:distri} obtained from Gowalla, a large location based social network (LBSN). As these figures show, there is a significant skew in terms of the number of users whose mobility data is collected and locations they check into. Further, the spatial distribution of mobility data is quite diverse as well. 
Therefore, current state-of-the-art methods struggle in low-data regions and the approaches that attempt to incorporate data from external sources suffer from the following limitations: \begin{itemize}

\item limited to cold-start users from within a city~\cite{onlycity, onlycity2, ruiruiwww}, \item focused only on using traffic network ignoring the use of social network of users and location dynamics~\cite{regiontrans, www, deepst}, \item generate trajectories using a model learned on traffic-network images of source region~\cite{similarwww, privacy3}, thus vulnerable to recaliberation, \item operating only for users who are \emph{common across} locations~\cite{mamo, compare}, or \item adopting a limited level of transfer through domain-invariant features~\cite{adit} --like the spending capacity or the users' age, thus constrained by the feature unavailability in public datasets.	\end{itemize}

% \sbcomment{do we need the following piece of sentence?} as well as susceptible to large fine-tuning.
Unfortunately, none of these approaches, especially those based on visual-data (i.e., traffic and location images), can be easily re-calibrated for a target POI network due to the varied spatial density, location category and lack of user-specific features in public POI datasets. 
%\sbcomment{we may want to elaborate this a bit?}
 
%We explore these variances in figures %We highlight the location-centric spatial variance in figures 1b and ~\ref{fig:c}, for the available POIs across two US states, California and Washington. Note that each region can be further classified into event spaces (e.g., university campus, business district etc.) which display potentially very different mobility dynamics. % of a subdivision based population, ex. university area, event spaces. 
%These data-specific constraints, privacy concerns and complex location dynamics and user preferences make our goal of transfer a non-trivial task. 

\subsection{Contributions}
In this paper, we present {\bfseries \ourm}(\textbf{A}utomated \textit{cross} \textbf{Lo}cation-network \textbf{T}ransfer \textbf{L}earning), a novel meta learning-based approach for POI recommendation in limited-data regions while transferring model parameters learned at a data-rich region \emph{without any prerequisite of inter-region user overlap}.
%Unlike previous approaches for handling data sparsity, we do not use a metapath-based procedure that requires tracking users between regions. 
%Unlike previous approaches, we don't require inter-region users, user traces between regions and is thus more suitable for transfer across geographically very distant regions (and even across different networks). 
%\ourm considers two transferable aspects in a mobility network, (a) general user-location preferences that include intra- and inter- user-location characteristics, and (b) locality specific transfer, i.e. preferences of a university district of the source to a similar district in target region. 
Specifically, we use a hierarchical multi-channel learning procedure with a novel meta-learning~\cite{maml, melu} extension for geographical scenarios called {\bf \textit{spatio-social} meta-learning (SSML)}, that learns the model parameters by jointly minimizing the region-specific social as well as location recommendation losses, and a cross-region transfer via \emph{clusters}~\cite{www, regiontrans} of user and locations with similar preferences by minimizing an alignment loss~\cite{what2tran, moreattn}, to achieve high performance even in extremely limited-data regions.
%, we minimize the divergence between user and location characteristics by minimizing an alignment loss\cite{what2tran, moreattn} based on \emph{clusters}~\cite{www, regiontrans} with similar preferences. 
We represent the POI network of each region via a heterogeneous graph with users and locations as nodes and capture the user-location inter-dependence and their neighborhood structure via a \emph{twin graph attention} model~\cite{danser, dualgcn}.
%\sbcomment{which characteristic? that of twin gats or axolotl transfer?} 
The graph-attention model aggregates all four aspects of user and location influences~\cite{socialgcn, eljp} namely, (i) users' social neighborhood and their location affinity to construct user-specific and location-conditioned representations, and (ii) similarly for each location, its neighboring locations and associated user affinity to obtain a \emph{locality-specific} and \emph{user-conditioned} representations. We combine this multi-faceted information to determine the \emph{final} user and location representations which are used for POI recommendation. We highlight the region-size invariant performance of \ourm by using regions with different spatial granularities, i.e. large and small \textit{states} for Germany and U.S respectively, and \textit{prefectures} for Japan. 

In summary, the key contributions we make in this paper via \ourm are three-fold:
\begin{enumerate}
\item \textbf{Region-wise Transfer:} We address the problems associated with POI recommendation in limited data regions and propose \ourm, a cross-region model transfer approach for POI recommendation that does not require common users and their traces across regions. %inter-region users and user traces between regions. 
It utilizes a novel \textit{spatio-social} meta-learning based transfer and minimizes the divergence between user-location clusters with similar characteristics.
%To accomplish this, we optimize an \textit{attentive} ~\cite{what2tran} based on user-location clusters. 
%\ourm is the first cross-region POI recommendation approach that neither \textit{tracks} users across regions and is impervious to regional . 
%This makes it more privacy conscious and suitable for learning across distant regions. 

\item \textbf{User-Location Influences:} Our twin graph attention-based model combines user and location influences in heterogeneous mobility graphs. 
%To the best of our knowledge, 
This is the first approach to combine these aspects for addressing the data scarcity problem with POI recommendation. \ourm is robust to network size, trajectory spread, and check-in category variance making it more suitable for transfer across geographically distant regions (and even across different networks). 

\item \textbf{Detailed Empirical Evaluation:}  We conduct thorough experiments over 12 real-world points-of-interest datasets from the U.S, Japan, and Germany, at different region-wise granularity. They highlight the superior recommendation performance of \ourm over state-of-the-art methods across all metrics. 
%Also, superior results across three real-world networks than all compared baselines, along with extensive research on parameter sensitivity supports the \ourm framework.
\end{enumerate}

\vspace{-0.1cm}
\section{Related Work}
\label{sec:relwork}
In this section, we introduce key related work for this paper. It mainly falls into following categories: 1) Mobility Prediction; 2) Graph based recommendations; 3) Transfer Learning and Mobility.\\

\noindent \textbf{Mobility Prediction:} 
Understanding the mobility dynamics of a user is widely studied using different data sources~\cite{zheng, cho}. Early efforts relied on taxi datasets to study individual trajectories~\cite{dcrnn, traffic}. However, these approaches are limited by the underlying datasets as it excludes two critical aspects of a mobility network; the social friendships and location categories. The social network is used to model the influence dynamics across different users~\cite{locate, cara} and the POI categories capture the different preferences of an individual~\cite{colab, cheng2013you}. We utilize user POI social networks for our model as these datasets provide both: a series of social dynamics for different users and location-specific interest patterns for a user. These are essential for tasks such as location-specific advertisements and personalized recommendations. Standard POI models that utilize an RNN~\cite{ganguly, cara, zheng, rnnlbsn, deepmove} or a temporal point process~\cite{imtpp, colab, reformd} are prone to irregularities in the trajectories. These irregularities arise due to uneven data distributions, missing check-ins, and social links. Moreover, these approaches consider the check-in trajectory for each user as a sequence of events and thus have limited power to capture the user-location inter-dependence through their spatial neighborhood, i.e. the location-sensitive information that influences all neighborhood events. Recent approaches~\cite{regiontrans, www}, harness the spatial characteristics by generating an image corresponding to each user trajectory and then utilize a CNN as an underlying model. Such approaches based on visual data, and all CNN-based approaches, are limited by the image characteristics such as resolution and interpolation. Modern POI recommendation approaches such as \cite{lbsn2vec} model the spatial network as a graph and utilize a random-walk-based model, with~\cite{ijcailbsn} proposing a graph-based neural network based model to incorporate structural information of the network. Unfortunately, none of these approaches are designed for mobility prediction in limited data regions. \\

\noindent \textbf {Graph based Recommendation:}
Existing graph embedding approaches focus on incorporating the node neighborhood proximity in a classical graph in their embedding learning process~\cite{node2vec, lbsn2vec}. For example ~\cite{birank} adopts a label propagation mechanism to capture the inter-node influence and hence the collaborative filtering effect. Later, it determines the most probable purchases for a user via her interacted items based on the structural similarity between the historical purchases and the new target item. However, these approaches perform inferior to model-based CF methods, since they do not optimize a recommendation-specific loss function.
The recently proposed graph convolutional networks (GCNs)~\cite{gcn} have shown significant prowess for recommendation tasks in user-item graphs. The attention-based variant of GCNs, graph attention networks (GATs)~\cite{gat} are used for recommender systems in information networks~\cite{socialgcn, ngcf}, traffic networks~\cite{traffic, dcrnn} and social networks~\cite{pinsage, yfumob}. Furthermore, the heterogeneous nature of these information networks comprises of multi-faceted influences that led to approaches with \textit{dual}-GCNs across both user and item domains~\cite{dualgcn, socialgcn}. However with POI networks, the disparate weights, location-category as node feature, and varied sizes, these models cannot be generalized for spatial graphs. \\

\noindent \textbf{Clustering in Spatial Datasets:}
In addition to the twin-GAT model, \ourm includes an alignment loss for cross-region transfer via clusters of users and locations with similar preferences. Due to the disparate features in our spatial graph, identifying the optimal number of clusters for the source and target region is a non-trivial task. Thus, we highlight a few key related works for clustering POIs and users in a spatial graph. Standard community-detection algorithms for spatial datasets~\cite{colab, com1, com2, com3} are not suitable for grouping POIs as they ignore the graph structure, POI-specific features such as categories, geographical distances, and the order of check-ins in a user trajectory. Moreover, the clustering performance of these methods is highly susceptible to the hyper-parameter values used in a setting. Recent approaches~\cite{morris2019weisfeiler, sagpool, diffpool, abu2019mixhop, gunet} can automatically identify the number of clusters in a graph by capturing higher-order semantics between graph nodes, however, their application to graphs in spatial and mobility domains have certain challenges. %Here, we highlight the probable application settings where these models may be used along with \ourm. 
In detail, (i) DiffPool~\citep{diffpool} can learn differentiable clusters for POIs and users for each region, however, these assignments are \textit{soft} \textit{i.e.,} without definite boundaries between distant POI clusters and, moreover, have a quadratic storage complexity; (ii)~\citet{gunet} ignores the topology of the underlying spatial graph; (iii)~\citet{sagpool} can be extended to spatial graphs, however, has limited scalability due to its self-attention~\cite{transformer} based procedure; (iv)~\citet{morris2019weisfeiler} can incorporate higher-order structure in a POI graph using multi-dimensional Weisfeiler-Leman graph isomorphism; and (v)~\citet{abu2019mixhop} can learn inter-user and inter-POI relationships by mixing feature representations of neighbors at various distances. However, due to the presence of two types of graph nodes -- user and POI -- identifying higher-order relationships by solely considering POI or user nodes is challenging. In addition, ~\citet{spat} uses a differentiable grouping network to discover the latent dependencies in a spatial network but is limited to air-quality forecasting. We highlight that though these approaches can be plugged-in with \ourm, they have an additional computation cost that gets further amplified due to the repetitive clustering procedure required in \ourm (please refer to Section~\ref{cluster_loss}). However, we note that using these approaches over \ourm while simultaneously maintaining the scalability is a probable future work of this paper.\\

\noindent \textbf{Transfer Learning and Mobility:}
Transfer learning has long been addressed for tasks involving sparse data~\cite{what2tran, maml} with applications to recommender systems as well~\cite{manasi, melu, compare}. Transfer-based spatial applications deploy CNNs across regions and achieve significant improvements in limited data-settings~\cite{regiontrans, www}. However, these approaches are restricted to non-structural data and a graph-based approach has not been explored by the previous literature. \cite{ruiruiwww} extends meta-learning to enhance recommendations in a POI setting, but is limited to a specific region. Information transfer across graphs is not a trivial task~\cite{transgraph, kgmaml, adit} and recent mobility models that incorporate graphs with meta-learning in~\cite{deepst, ltran} are either limited to traffic datasets and do not incorporate the social network or are limited to new trajectories~\cite{similarwww, privacy3}. From our experiments, we prove that a mere fine-tuning on the target data is susceptible to large cross-data variances and thus re-calibrating a generative model is not a trivial task in mobility-based networks.
\section{Problem Formulation} \label{sec:problem}
We consider POI data for two regions, a \textit{source} and a \textit{target} denoted by $\ds{D}^{src}$ and $\ds{D}^{tgt}$. We denote the users and locations in source and target networks as $\cm{U}^{src}, \cm{P}^{src} \in \ds{D}^{src}$ and $\cm{U}^{tgt}, \cm{P}^{tgt} \in \ds{D}^{tgt}$ correspondingly with no common entries $\cm{U}^{src} \cap \cm{U}^{tgt} = \cm{L}^{src} \cap \cm{L}^{tgt} = \varnothing$. In other words, we do not need a common user between two regions to perform a cross-region mobility knowledge transfer. With a slight abuse of notation, the network for any region --either target or source-- is assumed to consist of users $|\cm{U}| = M$, locations $|\cm{P}| = N$ and an affinity matrix $\bs{R}=\{r_{ul}\}_{M \times N}$. We populate entries in $\bs{R}$ as \textit{row-normalized} number of check-ins made by a user to a location (i.e., multiple check-ins mean higher value). This can be further weighed by the user-location ratings, if available. We denote a pair of users as $u_i, u_j$ and locations as $l_a, l_b$. We also assume that for each location $l_a$ we have one (or more) category label (such as Jazz Club, Cafe, etc.).
\begin{problem*}[\textbf{Target Region POI Recommendation}]
\label{checkin}
\textit{Given the mobility data of source and target regions, $\ds{D}^{src}$ and $\ds{D}^{tgt}$, our aim is to transfer the rich dynamics in source region to improve POI recommendation for users in the target region. Specifically, maximize the following probability:
\begin{equation}
P^* = \arg \max \{ \mathbb{E}[r_{u_i, l_a}^{tgt}| \ds{D}^{src}, \ds{D}^{tgt} ] \},
\end{equation}
where $\mathbb{E}[r^{tgt}_{u,l}]$ calculates the expectation of location $l_a$ in the target region being visited by the user $u_i$, thus $r_{u_i, l_a}^{tgt} \in \bs{R}^{tgt}$, given the mobility data of users from both source and target regions. Simultaneously for a region, our objective as personalized \emph{location} recommendation is to retrieve, for each user, a ranked list of candidate locations that are most likely to be visited by her based upon the past check-ins available in the training set.}
\end{problem*}

\subsection{User-Location Graph Construction}
For each region, we construct a heterogeneous user-location graph $\cm{G}^{src}$ and $\cm{G}^{tgt}$ but we describe generically as $\cm{G}=\{\cm{U}\cup\cm{P}, \cm{E}\}$ with each user and location as a node. The disparate edges $\cm{E}_u, \cm{E}_l, \cm{E}_r \in \cm{E}$ determine the user-user, location-location and user-location relationships respectively. The structure of the graph is as follows: 
\begin{asparaitem} 
\item An edge, $e_{u_i,u_j} \in \cm{E}_u$, between two users, $u_i$ and $u_j$ is denotes a social network friendship.
\item We form an edge, $e_{l_a, l_b} \in \cm{E}_l$, between two locations when any user has \emph{consecutive check-ins} between them -- i.e., a check in at $l_a$ (or $l_b$) followed immediately by $l_b$ (or $l_a$) with edge weight based on the \emph{geographical distance}~\cite{dcrnn, yfumob} between the two locations. Specifically, we use non-linear decay with distance as:
\begin{equation*}
w (e_{l_a, l_b}) = 
\begin{cases}
\exp \left(-\frac{d(l_a, l_b)}{\sigma^2}\right), & \mathrm{if~} d(l_a, l_b) \le \kappa, \\
0 & \mathrm{otherwise},
\end{cases}
\end{equation*}
where $d(l_a, l_b)$ is the \emph{haversine}~\cite{haversine} distance between the two locations.
%where $d(l_a, l_b)$ is the \emph{haversine} distance\footnote{\href{https://en.wikipedia.org/wiki/Haversine\_formula}{https://en.wikipedia.org/wiki/Haversine\_formula}} between the two locations. 
%given by: $$ d(l_a, l_b) = 2r \arcsin \sqrt{\sin^2\left ( \frac{\Delta \psi}{2} + \cos\psi_a \cos\psi_b \sin^2\left (\frac{\Delta\lambda}{2} \right )\right)},$$ where $\psi, \lambda$ denote the co-ordinates of $l_a$ and $l_b$ respectively.
%Unlike previous approaches~\cite{ijcailbsn} that use visit frequency, this weighting scheme is consistent across source and targets, hence is less-conditioned on the region-specific data and enforces a \emph{location} awareness for each node.
\item A check-in by user $u_i$ at location $l_a$ results in a user-location edge, $e_{u_i,l_a} \in \cm{E}_r$.
\end{asparaitem}
%In Figure~\ref{fig:arch}, we have colored user nodes of our heterogeneous graph with yellow, and location nodes with blue. 
\noindent We also use the following notations to define different neighborhoods that will be used in our model description later: 
\begin{inparaenum}
\item{$\mathcal{N}_{u_i} = \left\{u_k : e_{u_i,u_k} \in \mathcal{E}_u\right\}$: the social neighborhood of user $u_i$,}
\item{$\mathcal{N}_{l_a} = \left\{u_k : e_{u_k,l_a} \in \mathcal{E}_r\right\}$: the user neighborhood of location $l_a$,}
\item{$\mathcal{S}_{u_i} = \left\{l_k : e_{u_i, l_k} \in \mathcal{E}_r\right\}$: the location neighborhood of user $u_i$, and, finally,}
\item{$\mathcal{S}_{l_a} = \left\{l_k : e_{l_a, l_k} \in \mathcal{E}_l\right\}$: locations in the spatial vicinity of $l_a$.}
\end{inparaenum}
Note that the graph $\mathcal{G}$ can also be enriched with all edges as weighted~\cite{ijcailbsn} conditioned on the availability of different features. However, we present a general framework which can be easily extended to such settings. %We learn different the influences in the graph using corresponding graph attention networks, which form the basis for the \ourm framework.

\begin{table}[t!]
\caption{\label{tab:par} Summary of Notations Used.}
\vspace{-0.4cm}
%\resizebox{0.7\columnwidth}{!}{
\centering
\begin{tabular}{ll}
\toprule
\textbf{Notation} & \textbf{Description}\\
\midrule
$\ds{D}^{src}, \ds{D}^{tgt}$ & Datasets for source and target regions\\
%$\cm{U}, \cm{L}$ & Set of all user and locations for a region\\
$\bs{R}$ & Normalized user to location preference matrix\\
$\bs{U}_l, \bs{U}_s, \bs{U}_f$ & User's latent, location-based and final representations\\
$\bs{L}_l, \bs{L}_s, \bs{L}_f$ & Location's latent, user-based and final representations\\
$\Phi_{1\cdots4}$ & Graph Attention Networks in \ourm\\
%$\bs{W}_{\Phi_{1\cdots4}}, \bs{b}_{\Phi_{1\cdots4}}$ & Weight matrix and bias vector for $\Phi_{1\cdots4}$\\
%$\bs{W}_{1\cdots4}, \bs{b}_{1\cdots4}$ & Weight matrix and bias vector for $\Phi_{1\cdots4}$\\
%$\bs{\alpha}^{\Phi_1\cdots\Phi_4}$ & Attention weights matrices for $\Phi_{1\cdots4}$\\
$\psi_{1\cdots4}$ & MLP layers to calculate attention weights for $\Phi_{1\cdots4}$\\
$\varphi_{1\cdots3}$ & MLPs for final user and location embeddings, and affinity prediction\\
%$\bs{G}_{1\cdots4}$ & Weight matrix for attention function $\phi_{1\cdots4}$\\
$\cm{L}_{p}, \cm{L}_{s}, \cm{L}_{c}$ & Affinity, social prediction and Cluster-based loss\\
%$D$ & Embedding dimension for users and locations \\
$K$ & No. of clusters in source and target region\\
$N_u$ & No. of target region based updates for SSML\\
$M_t$ & Iterations before cluster-transfer\\
$\bs{U}^{tgt}_c, \bs{L}^{tgt}_c$ & User and location clusters for target region\\
$\bs{U}^{src}_c, \bs{L}^{src}_c$ & User and location clusters for source region\\
%$\bs{\beta}^{u,l}$ & Attention matrices for cluster-alignment loss $\cm{L}_c$\\
\bottomrule
\end{tabular}
%}
\end{table}

\section{\ourm Framework}
\label{sec:framework}
In this section we first describe in detail the basic model of \ourm along with its training procedure. Then we present the key feature of \ourm, {\it viz.}, its ability to transfer model parameters learned from data-rich region to data-scarce region. For a specific region, we embed all users in $\cm{U}$ and all locations in $\cm{P}$ through embedding matrices $\bs{U}=\{\bs{u}_{i}\}_{M \times D}$ and $\bs{L}=\{\bs{l}_{a}\}_{N \times D}$ respectively with $D$ as the embedding dimension. A summary of all notations is given in Table~\ref{tab:par}. %We will discuss the information transfer in \ourm by meta-learning in the next section. %In this section we elaborately describe construction methodology for the LBSN graph, our proposed model \ourm along with its training procedure for information transfer across different regions. 

\subsection{Basic Model}
\begin{figure}[t] 
\centering
  \includegraphics[width=0.6\linewidth]{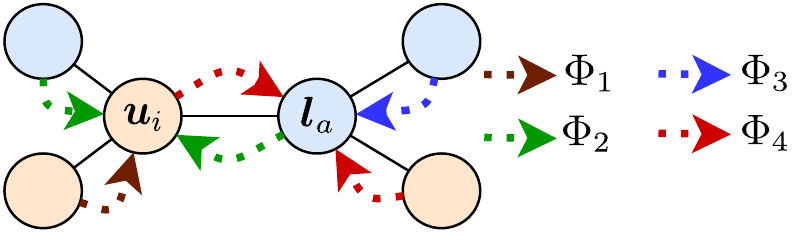}
  \vspace{-4mm}
  \caption{Different node and edge types in our graph model (users are in yellow and locations are in blue). The dashed arrows represent various influences and corresponding GATs ($\Phi_i, i=1,2,3,4$) in \ourm.}
\label{fig:social}
\vspace{-2mm}
\end{figure}

%\sbcomment{do we need the following line?}Earlier works have shown that users' \textit{latent preference} is a complex mixture of the influence from their social network and their own preference-based behavior, which are difficult to capture using simple MLP or GCN-based models, especially when combined with location preferences~\cite{eljp, socialgcn, ijcailbsn}. %This affinity along with location preferences exhibits complex characteristics that are difficult to capture by a simple MLP or GCN based model~\cite{socialgcn, ijcailbsn}. 
In the graph model of \ourm, called \ours-basic, we capture the four aspects of influence propagation --namely, user-latent embeddings ($\bs{U}_l \in \mathbb{R}^{M \times D}$), location-conditioned user embeddings ($\bs{U}_s\in \mathbb{R}^{M \times D}$), location-latent embeddings ($\bs{L}_l \in \mathbb{R}^{N \times D}$), and user-conditioned location embeddings ($\bs{L}_s \in \mathbb{R}^{N \times D}$)-- illustrated in Figure ~\ref{fig:social}. 

\begin{figure}[t!] 
\centering
  \includegraphics[width=0.7\linewidth]{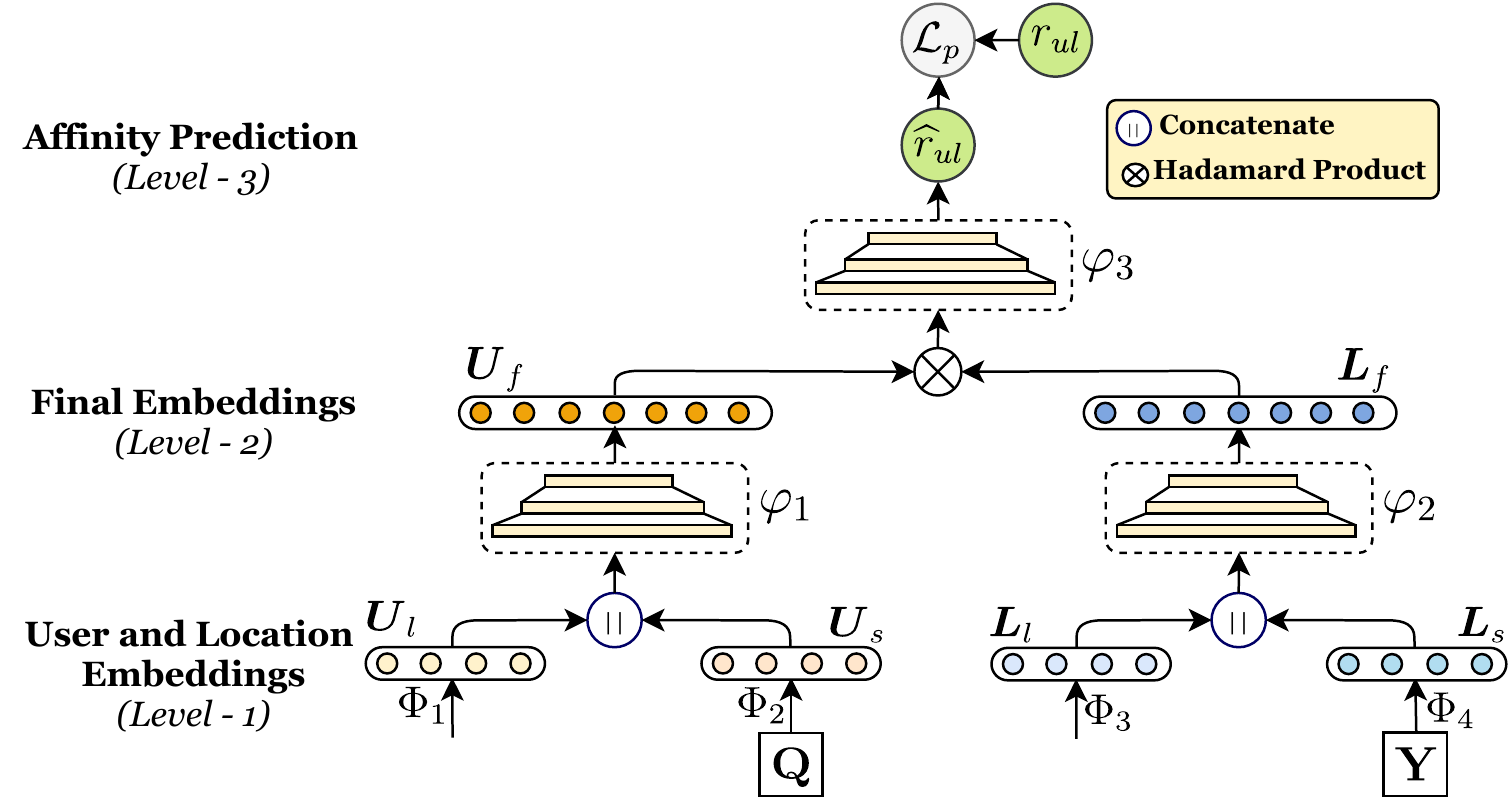}
\vspace{-4mm}  
  \caption{System architecture of \ours-basic that shows level-wise embedding computation and affinity prediction. User-Latent ($\bs{U}_l$) and Location conditioned ($\bs{U}_s$) embeddings are combined to form a \textit{final} user embedding ($\bs{U}_f$) and similarly for location, $\bs{L}_l$ and $\bs{L}_s$ are combined to $\bs{L}_f$.}
\label{fig:arch}
\vspace{-2mm}
\end{figure}

The basic model of \ourm captures these four aspects using different graph attention networks, resulting in a twin-graph architecture as shown in  Figure~\ref{fig:arch}. In the rest of this section, we first describe each of the graph attention components and the prediction model in the basic model of \ourm. Subsequently, we delineate the information transfer component that operates over this basic model.

%These representations are learned each using a different graph attention network.\\ \\
\noindent\textbf{GAT for User Latent Embedding($\Phi_1$):}
In each iteration, using the available user embeddings, $\bs{U}$, we aggregate each user's social neighborhood to obtain a new \emph{latent} representation of the user. 
%Using the available user embedding, $\bm{U}$, we aggregate a user-node's local neighbors to get the new \emph{latent} representation of the user. 
We denote this embedding as $\bs{U}_l$ and calculate as follows:
\begin{equation} 
\bs{u}_{l, i} = \sigma  \bigg( \sum_{u_k \in \cm{N}_{u_i}} \bs{\alpha}^{\Phi_1}_{u_i, u_k}\left( \bm{W}_{\Phi_1} \bs{u}_k + \bm{b}_{\Phi_1} \right)  \bigg), \, \bs{u}_{l,i}\in \bs{U}_l,
\end{equation}
where $ u_i \in \cm{U}, \cm{N}_{u_i},\sigma, \bm{W}_{\Phi_1}$ and $\bm{b}_{\Phi_1}$ are the user node ${u}_i$, nodes in the social neighborhood of ${u}_i$, the activation function, the weight matrix and the bias vector respectively. $\bs{\alpha}^{\Phi_1}_{u_i, u_k}$ determines the attention weights between user embeddings $\bs{u}_k$ and $\bs{u}_i$ %in its neighborhood, 
given by: % and $*$ denotes embedding for the \textit{current} iteration.
\begin{equation}
\label{eqn:alpha}
\bs{\alpha}^{\Phi_1}_{u_i, u_k} = \frac{\exp \big(\psi_1(\bs{u}_i, \bs{u}_k) \big)}{\sum_{u_j \in \cm{N}_{u_i}} \exp \big(\psi_1(\bs{u}_i, \bs{u}_j) \big)},
\end{equation}
where, $\psi_1(\bs{u}_i, \bs{u}_j) = \mathtt{LeakyReLU} (\bm{G}_{\Phi_1} \otimes(\bs{u}_i \mathbin\Vert \bs{u}_j))$ calculates the inter-user attention weights with learnable parameter $\bm{G}_{\Phi_1}$.

\noindent\textbf{GAT for Location-conditioned User Embeddings ($\Phi_2$):} To encapsulate the influence on a user based on the her check-ins as well as those by her social neighborhood, our embeddings must include location information for all check-ins made by different users in her social proximity. For this purpose, we first need to aggregate the location embeddings for every check-in made by a user ($u_i$) as her location-based embedding, $\bs{Q} = \{ \bs{q}_i \}_{M \times D}$. We note that the category of a check-in location is arguably the root-cause for a user to visit the location and thus to capture the location-specific category and the user-category affinity in these embeddings, we populate $\bs{Q}$ using a max-pooling aggregator across each location embedding weighted by the probability of a category to be in a user's check-in locations. That is, 
%
%Earlier approaches~\cite{delf, danser} simply aggregated these representations based on attention weights alone. However, it ignores the location-specific category (e.g., Jazz Club, Cafe, etc.) which arguably is the \emph{raison d'\^etre} for user to visit the location. We use a \emph{max-pool} across each location embedding weighted by the probability of occurrence of the category in the user's check-ins to succinctly capture the category information. That is,
%
%Earlier approaches aggregated these representations based on attention weights. However, this approach cannot be incorporated in LBSN as it ignores location-specific category e.g., Jazz Club, Cafe, which is the foremost important feature of a location for attracting a user. To reflect the location features and parallelly reduce the no. of training parameters, we consider a \emph{max-pool} across each location embedding weighted by the probability of category occurrence in the user's check-ins. 
\begin{equation}
\bs{q}_{i} = \mathtt{MaxPool} \bigg [ \sum_{l_k \in \cm{S}_{u_i}} \mathrm{p}^{u_i}_{(l_k)} \cdot \bs{l}_k \bigg], \quad \forall \bs{q}_{i} \in \bs{Q},
\end{equation}
where $\cm{S}_{u_i}$, $\mathrm{p}^{u_i}_{(l_k)}$ respectively denote the location neighborhood of $u_i$ and the probability of a POI-category to be present in the past check-ins of $u_i$. We calculate $\mathrm{p}^{u_i}_{(l_k)}$ as the fraction of check-ins made by the user to POIs of the specific category with the total number of her check-ins. Mathematically,
\begin{equation}
\mathrm{p}^{u_i}_{(l_k)} = \frac{\text{No. of check-ins by $u_i$ at POIs with category same as $l_k$}}{\text{Total no. of check-ins by $u_i$}},
\end{equation}
We calculate these probabilities for every POI category and these values are user-specific. Moreover, the values of $\mathrm{p}^{u_i}_{(l_k)}$ can be considered as the explicit category preferences of a user $u_i$.\\
Later, to get the influence of neighborhood locations on a user, we aggregate the location-based neighbor embeddings $\bs{Q}$ for each user (${u}_i$) based on her social network as:
\begin{equation}
\bs{u}_{s,i} = \sigma \bigg( \sum_{u_k \in \cm{N}_{u_i}} \bs{\alpha}^{\Phi_2}_{u_i, u_k}\left( \bm{W}_{\Phi_2} \bs{q}_k + \bm{b}_{\Phi_2}\right) \bigg), \, \bs{u}_{s, i}\in \bs{U}_s,
\end{equation}
where $\bs{\alpha}^{\Phi_2}_{u_i, u_k}$ is the attention weight for quantifying the influence a user has on another through its check-ins and is formulated using $\psi_2$ similar to $\psi_1$ (Eqn ~\ref{eqn:alpha}). %$\bs{\alpha}^{\Phi_1}$ and $\phi_1$ in eqn \ref{eqn:alpha}. 
The resulting embedding $\bs{U}_s$ is the location-conditioned user embedding.
%$$\alpha^q_{ij} = \frac{\exp(\phi_2(\mathbf{q}_i, \textbf{q}_j))}{\sum_{k \in \mathcal{N}^u_i} \exp(\phi_2(\textbf{q}_i, \textbf{q}_k))},
%$$
%where, $\phi_2(\cdot)$ is similar to $\phi_1(\cdot)$ with weight parameter $\mathbf{W}_2$.

\noindent\textbf{GAT for Location Latent Embedding ($\Phi_3$):} Similar to $\Phi_1$, we aggregate the neighborhood of each location, $\bs{l}_a \in \bs{L}$, to get a \textit{latent} representation of each location. %However, each associated edge between locations is weighted by its distance. 
To factor the inter-location edge-weight in our embeddings, we sample locations from the neighborhood with probability proportional to $w(e_{l_a, l_b})$, i.e., closer the locations higher their repetitive sampling. These sampled locations represent the vicinity of the check-in and we encapsulate them to get the latent location representation.
\begin{equation}
\bs{l}_{l, a} = \sigma \bigg( \sum_{l_k \in \cm{S}_{l_a}} \bs{\alpha}^{\Phi_3}_{l_a, l_k}\left( \bm{W}_{\Phi_3} \bs{l}_k + \bm{b}_{\Phi_3} \right) \bigg), \, \forall l_a \in \cm{P}, \bs{l}_{l,a}\in \bs{L}_l,
\end{equation}
where $\bs{\alpha}^{\Phi_3}_{l_a, l_k}$ is again formulated as in Equation \ref{eqn:alpha}, using $\psi_3(\bs{l}_a, \bs{l}_k)$.

\noindent\textbf{GAT for User-conditioned Location Embedding($\Phi_4$):} Similar to $\Phi_2$, we need to capture the influence of different users with check-ins nearby to the current location $l_a$. Thus we use a \emph{max-pool} aggregator to capture the user neighborhood of each location weighted by its affinity towards the location category. Through this, we aim to encapsulate the locality-specific user preferences, i.e. the \textit{counter}-influence of $\bs{Q}$ in $\Phi_2$, denoted as $\bs{Y} = \{\bs{y}_a\}_{N \times D}$. 
\begin{equation}
\bs{y}_{a} = \mathtt{MaxPool} \bigg[ \sum_{{u}_k \in \cm{N}_{l_a}} \mathrm{p}^{l_a}_{(u_k)} \cdot \bs{u}_k \bigg], \quad \forall \bs{y}_a \in \bs{Y}, 
\end{equation}
where, $\mathrm{p}^{l_a}_{(u_k)}$ denotes the category affinity of all users in the neighborhood $\cm{N}_{l_a}$ of a location $l_a$. We calculate $\mathrm{p}^{l_a}_{(u_k)}$ for each user as the fraction of check-ins of a user $u_k$ with the total check-ins for all users in $\cm{N}_{l_a}$ at the POIs with category same as $l_a$. Mathematically,
\begin{equation}
\mathrm{p}^{l_a}_{(u_k)} = \frac{\text{No. of check-ins by $u_k$ with category \textit{`cat'}}}{\text{No. of check-ins by users in $\cm{N}_{l_a}$ with category \textit{`cat'}}},
\end{equation}
where \textit{`cat'} denotes the category of POI $l_a$. Moreover, these probabilities are specific to each POI and can be interpreted as the affinity of nearby users towards the POI category. Similar to $\Phi_3$, we aggregate the location neighborhood using an edge-weight based sampling on $\bs{Y}$ to get user-conditioned location embedding $\bs{L}_s$ with parameters $\bm{W}_{\Phi_4}, \bm{b}_{\Phi_4}$ and $\bs{\alpha}^{\Phi_4}$ and $\psi_4(\bs{y}_a, \bs{y}_b)$. 

\subsection{Model Prediction}\label{basic_predict}
We combine the four representations of users and locations developed above using fully-connected layers, with a concatenated input of $\bs{U}_l$ and $\bs{U}_s$ for final user embedding $\bs{U}_f = \varphi_1\left(\bs{U}_l \mathbin\Vert \bs{U}_s\right)$; and similarly for locations $\bs{L}_l$ and $\bs{L}_s$ to obtain %for \textit{final} user embedding   and \textit{final} 
final location embedding $\bs{L}_f = \varphi_2\left(\bs{L}_l \mathbin\Vert \bs{L}_s\right)$. Finally, we estimate a user's affinity to check-in at a location by a Hadamard (element-wise) product between the corresponding representations. Formally, 
\begin{equation}
\widehat{r_{ul}} = \varphi_3\left(\bs{U}_f \, \otimes \, \bs{L}_f\right),
\end{equation}
where $\varphi_1(\cdot)$, $\varphi_2(\cdot)$ and $\varphi_3(\cdot)$ represent fully-connected neural layers.

The parameters are optimized using a \textit{mean-squared} error that considers the difference between the user's predicted and the actual affinity towards a location, with $L_1$ regularization over the trainable parameters.
\begin{equation}
\cm{L}_{p} =  \sum_{(\forall u, l)} \snorm{\widehat{r_{ul}} - r_{ul}}^2 + \lambda_{p}\snorm{\Theta_{pred}}.
\end{equation}
$\Theta_{pred}$ refers to all the trainable parameters in \ourm for a region-specific prediction including the weights for all attention networks. 

\subsection{\ourm: Information Transfer}\label{inftran}
We now turn our attention to the central theme of this paper, namely, the training procedure for \ourm along with its cluster-wise transfer approach. We reiterate that we do not expect any common users/POIs between source and target regions, and thus the only feasible way to transfer mobility knowledge using the trained model parameters and the user-POI embeddings.  Specifically, there are two channels of learning for \ourm, (1) Spatio-Social Meta-learning based optimization, and (ii) Region-wise cluster alignment loss.

\subsubsection{Spatio-Social Meta-Learning (SSML)}\label{meta_part}
\begin{figure}[b!]
\begin{subfigure}[]{0.32\columnwidth}
\centering
%\vspace{1mm}
\includegraphics[width=\linewidth]{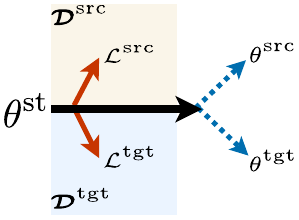}
%\vspace{1mm}
\caption{MAML}
\label{fig:a}
\end{subfigure}
\hspace{0.3cm}
\begin{subfigure}[]{0.50\columnwidth}
\centering
\includegraphics[width=\linewidth]{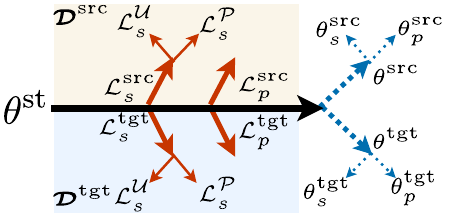}
\caption{SSML}
\label{fig:b}
\end{subfigure}
\vspace{-4mm}
\caption{The difference between MAML(fig~\ref{fig:a}) and our SSML (fig~\ref{fig:b}) is that the latter optimizes parameters in a hierarchy i.e. user and location parameters are updated for neighborhood prediction and then combined for POI recommendation. (Best viewed in color).}
\label{fig:maml}
\end{figure}
Meta-learning has long been proposed to alleviate data scarcity problem in spatial datasets~\cite{deepst, www, ruiruiwww}. Specifically, in meta-learning, we aim to learn a joint parameter initialization for multiple tasks by simultaneously optimizing the prediction loss for each task. However, there is a high variance in data-quality between the regions $\ds{D}^{src}$ and $\ds{D}^{tgt}$ and thus a \textit{vanilla} meta-learning ---also called a model agnostic meta-learning (MAML)~\cite{maml}--- will not be sufficient as the target region is expected to have nodes with limited interactions. Such nodes, due to their low contribution in the loss function, may get neglected during the meta-procedure. We overcome this via a \textit{hierarchical} learning procedure that not only considers the location recommendation performance, but also the social neighborhood of all user and location nodes. We call the resulting learning procedure as \textbf{spatio-social meta-learning} (SSML) and the contrast between this approach and standard model agnostic meta-learning approach (MAML)~\cite{maml} is schematically given in Figure ~\ref{fig:maml}. 

Specifically, we consider the two tasks of \begin{inparaenum}[(i)] \item optimizing the POI recommendation loss function $\cm{L}_{p}$ across both source and target regions, and, \item neighborhood prediction for each node in both the networks~\cite{hsml}.\end{inparaenum} We initialize the parameters for the recommender system with global initial values ($\theta^{st}$) shared across both source and target. Note that by $\theta^{st}$ we mean the parameters for all GATs ($\Phi_{1\cdots4}$) and prediction MLPs ($\varphi_{1\cdots3}$) and thus exclude region-specific, user and location embedding matrices, $\bs{U}^{src}, \bs{L}^{src}, \bs{U}^{tgt}$ and $\bs{L}^{tgt}$.
We describe the hierarchical learning procedure of SSML here:

\noindent \underline{\textit{Neighborhood Prediction}}: For any region target or source, consider a user $u_i$ and her neighbor $u_j \in \cm{N}_{u_i}$, we obtain the probability of them being connected on the social network as $\hat{v}_{u_i, u_j} = \bs{u}_i \cdot \bs{u}_j^T$ where $\bs{u_\bullet} \in \bs{U}_f$ represents the \textit{final} representations of a user. We optimize the following cross-entropy loss:
\begin{equation}
\cm{L}^{\cm{U}}_{s}(\ds{D}^{\bullet}) = -\sum_{u_i \in \cm{U}^{\bullet}} \sum_{\substack{u_j \in \cm{N}_{u_i}\\u'_j \notin \cm{N}_{u_i}}} \left[ \log \left ( \sigma(\hat{v}_{u_i, u_j}) \right) + \log \left (1 - \sigma(\hat{v}_{u_i, u'_j}) \right) \right],
\end{equation}
where, $\hat{v}_{u_i, u_j}, \hat{v}_{u_i, u'_j}, \sigma$ denote the estimated link probability between two users connected by their social networks, with a negatively sampled user $u'_j$ \textit{i.e.} a user not in $\cm{N}_{u_i}$, and the \textit{sigmoid} function.
Similarly, we calculate the probability of a location $l_b$ being in the spatial neighborhood of a location $l_a$ as $\hat{v}_{l_a, l_b} = \bs{l}_a \cdot \bs{l}_b^T$ and denote the neighborhood loss as $\cm{L}^{\cm{P}}_{s}(\ds{D}^{\bullet})$. For the region as a whole, say target, the net neighborhood loss is defined as:
\begin{equation}
\cm{L}^{tgt}_{s} = \cm{L}^{\cm{U}}_{s}(\ds{D}^{tgt}) + \cm{L}^{\cm{P}}_{s}(\ds{D}^{tgt}),
\end{equation}
Similarly for source regions we denote social loss as $\cm{L}^{src}_{s}$.

\noindent \underline{\textit{POI Recommendation}}: %As mentioned in \ref{basic_predict}, our recommendation loss for source and target regions is denoted by $\cm{L}^{src}_{p}$ and $\cm{L}^{tgt}_{p}$ respectively. 
Since \ourm is designed for limited data regions, we purposely incline the meta-procedure towards improved target-region predictions. Specifically, we alter the meta-learning procedure by optimizing the parameters with the recommendation loss for the target region ($\cm{L}^{tgt}_p$) for a pre-defined no. of updates ($N_{u}$) and then optimize for the source region prediction loss($\cm{L}^{src}_p$).
\begin{align}
\nonumber \mathrm{Target:} & \, \theta^{tgt}_{k+1} \leftarrow 
\begin{cases} 
	\theta^{tgt}_k - \omega_1 \nabla_{\theta^{tgt}_k}\,\cm{L}^{tgt}_{p}(f_{\theta^{tgt}_k}), & \forall \, 1 \le k \le N_u, \\
	\theta^{st} - \omega_1 \nabla_{\theta^{st}}\,\cm{L}^{tgt}_{p}(f_{\theta^{st}}), & \mathrm{otherwise}, 
\end{cases}\\
 \mathrm{Source:} & \, \theta^{src} \leftarrow \theta^{st} - \omega_1 \nabla_{\theta^{st}}\,\cm{L}^{src}_{p}(f_{\theta^{st}}),
\end{align}
where $\theta^{st}, \theta^{tgt}, \theta^{src}, \cm{L}^{tgt}_{p}, \cm{L}^{src}_{p}, f_\theta$ are the global model independent parameters, parameters for target and source region, prediction loss for target and source, and \ourm output respectively. 

\noindent \underline{\textit{Final Update:}} The final update to global parameters is done by: (i) optimizing the region-specific social loss, $\cm{L}^{tgt}_{s}$ and $\cm{L}^{src}_{s}$, and (ii) minimizing the POI recommendation loss $\cm{L}^{tgt}_{p}, \cm{L}^{src}_{p}$ for both source and target regions.
\begin{equation}
\theta^{st} \leftarrow \theta^{st} - \omega_2 \cdot \left[ \nabla_{\theta^{st}}\,\cm{L}^{src}_{p}(f_{\theta^{st}}) + \nabla_{\theta^{tgt}_{N_u}}\,\cm{L}^{tgt}_{p}(f_{\theta^{tgt}_{N_u}}) \right] - \omega_3 \cdot \left[ \nabla_{\theta^{st}}\,\cm{L}^{src}_{s}(f_{\theta^{st}}) + \nabla_{\theta^{st}}\,\cm{L}^{tgt}_{s}(f_{\theta^{st}}) \right],
\label{eqn:update}
\end{equation}
where, $\omega_2, \omega_3$ denote the region-wise learning rates. 
%\begin{equation}
%\theta_{st} \leftarrow \theta_{st} - \omega_2 \cdot \left[\nabla_{\theta_{st}}\,\cm{L}^{src}_{p}(f_{\theta_{st}}) + \nabla_{\theta_{tgt}^{N_u}}\,\cm{L}^{tgt}_{p}(f_{\theta_{tgt}^{N_u}}) + \nabla_{\theta_{st}}\,\cm{L}^{soc}_{s}(f_{\theta_{st}}) + \nabla_{\theta_{st}}\,\cm{L}^{tgt}_{s}(f_{\theta_{st}}) \right]
%\end{equation}
%For our experiments, we suitably weight the social prediction loss using a hyper-parameter $\gamma$ and also show the benefits of this training procedure. 
%We refer this learning procedure as \textit{Spatial-social} meta-learning (SSML) and the contrast between this approach and MAML is given in Figure ~\ref{fig:maml}. 
%This variant of our model that uses only the SSML based optimization is referred to as \textbf{\ours-m}. %The convergence analysis of such a learning is provided in~\cite{hsml}.

\begin{algorithm}[t]
\DontPrintSemicolon
\KwIn{\\$\ds{D}^{src}$: Source-Region Training Data, $\ds{D}^{tgt}$: Target-Region Training Data \\
$M_t$: Epoch-based checkpoint, $\cm{C}$: Clustering function \\
}
\KwOut{$\theta^{tgt}$: Trained \ourm Parameters for Target Region}
$\theta^{st}, \theta^{tgt}, \theta^{src} \leftarrow$ Randomly initialize all parameters\\
\SetKwBlock{Begin}{function}{end function}
  \While {$\mathtt{epoch} < \mathtt{Max\_Epoch}$}
  {
  	Parameter update via target-region social prediction: $\theta^{tgt}_{0} \leftarrow \theta^{st} - \omega \nabla_{\theta^{st}}\,\cm{L}^{tgt}_{s}(f_{\theta^{st}})$\\
    Parameter update via source-region social prediction: $\theta^{src}_{0} \leftarrow \theta^{st} - \omega \nabla_{\theta^{st}}\,\cm{L}^{src}_{s}(f_{\theta^{st}})$\\
    
  	Calculate the target-region recommendation loss: $\cm{L}^{tgt}_p \leftarrow \mathtt{PredictionLoss}(\ds{D}^{tgt})$\\
  	Initial update-before iterations: $\theta^{tgt}_{1} \leftarrow \theta^{st} - \omega_1 \nabla_{\theta^{st}}\,\cm{L}^{tgt}_{p}(f_{\theta^{st}})$\\
    \For{$k < N_u$}
    { 
    	Iterative updates: $\theta^{tgt}_{k+1} \leftarrow \theta^{tgt}_k - \omega_1 \nabla_{\theta^{tgt}_k}\,\cm{L}^{tgt}_{p}(g_{\theta^{tgt}_k})$\\
    }
    Calculate the source-region recommendation loss: $\cm{L}^{src}_p \leftarrow \mathtt{PredictionLoss}(\ds{D}^{src})$\\
    Update for source parameters $\theta^{src} \leftarrow \theta^{st} - \omega_1 \nabla_{\theta^{st}}\,\cm{L}^{src}_{p}(f_{\theta^{st}}(\cm{R}_t))$\\
    
    Joint update for global parameters: As in eqn. \ref{eqn:update} \\
    \uIf{$\mathtt{epoch}\mod M_t$}
	{
		$\cm{L}_c \leftarrow \mathtt{ClusterLoss}(\cm{C}, \ds{D}^{src}, \ds{D}^{src})$\\
		Initialize cluster-based transfer $\theta^{src},\theta^{tgt} \leftarrow \displaystyle\min \cm{L}_c$
	}
	$\mathtt{epoch}++$\\
  }\label{endfor}
  Fine-tune for target region: $\theta^{tgt} \leftarrow \texttt{FineTune}(\cm{L}^{tgt}_{p})$ \\
  Return model parameters: \textbf{return} $\theta^{tgt}$
\caption{Training Algorithm for \ourm}\label{axoalgo}
\end{algorithm}

\subsubsection{Cluster Alignment Loss}\label{cluster_loss}
Recent research~\cite{www, regiontrans} has shown that enforcing similar patterns across specific POI clusters between source and target domains, e.g. from one university campus to another facilitates better knowledge transfer. Unfortunately, obtaining the necessary semantic information to align similar clusters across regions, that these techniques require, is not always practical at large-scale settings. For a POI network, a rudimentary approach to identify clusters would be to traverse across the categories associated with each location -- which may be quite expensive to compute and will neglect the user-dynamics as well.
%We avoid these approaches and instead rely on the robust graph-attentive model of \ourm to capture the user- and location-attributes that are optimized to be \textit{latent} as well \textit{conditioned} on one-another. 
We avoid these approaches and use a \textit{light-weight} Euclidean-distance based \textit{k-means} clustering to identify a set of users and locations in source as well as target regions (separately) that have displayed \textit{similar} characteristics till the current iteration. Such a dynamic clustering mechanism over the contemporary GAT embeddings prevents the need for additional hand-crafting. In contrast to previous approaches~\cite{hsml, locate}, we utilize a \textit{hard}-assignment as unlike \textit{online} product purchases, the mobility of a user is bounded by geographical distance~\cite{cho} and thus checkins to distant locations are very unlikely.
%. For each location cluster, we consider all the users with at least one visit to any location in the cluster as a user-cluster with affinity to that location cluster. 
We denote $\bs{U}^{tgt}_c$, $\bs{L}^{tgt}_c$, $\bs{U}^{src}_c$, $\bs{L}^{src}_c \in \mathbb{R}^{K\times D}$ and $\cm{C}$ as the cluster embedding matrices for target region-users, locations, source region-users, locations and the clustering algorithm respectively. We calculate the cluster embedding by \textit{mean-pooling} the embeddings of elements in the cluster. For example, we calculate embeddings for user-clusters in the target region as:
\begin{equation}
\bs{u}^{tgt}_{c} = \mathtt{MeanPool}\left[\bs{u}_i \cdot \Psi(u_i, c)\right]  \, \forall u_i \in \cm{U}^{tgt}, \bs{u}^{tgt}_{c} \in \bs{U}^{tgt}_{c},
\end{equation}
where, $\Psi(u_i, c)$ is the indicator function denoting whether user $u_i$ belongs to cluster $c$. Similarly we calculate $\bs{L}^{tgt}_c$, $\bs{U}^{src}_c$ and $\bs{L}^{src}_c$. For minimizing the divergence between the similar users and POIs between across regions, instead of engineering an explicit alignment between clusters, we use an attention-based approach to identify and align clusters with similar patterns and minimize the corresponding \textit{weighted} $L_2$ loss~\cite{moreattn, what2tran}.
\begin{equation}
\cm{L}_c =  \sum_{\forall c^t_u \in \bs{U}^{tgt}_c} \sum_{\forall c^t_l \in \bs{L}^{tgt}_c} \norm{\bs{c}^t_u - \sum_{c^s_u \in \bs{U}^{src}_c}\bs{\beta}^u_{c^t_u, c^s_u} \bs{c}^s_u}^2 + \, \norm{\bs{c}^t_l - \sum_{c^s_l \in \bs{L}^{src}_c}\bs{\beta}^l_{c^t_l, c^s_l} \bs{c}^s_l}^2,
\end{equation}
where $\bs{\beta}^u, \bs{\beta}^l \in \mathbb{R}^{K \times K}$, are the attention matrices for user and location clusters respectively. Each index in $\bs{\beta}^u, \bs{\beta}^l$ denotes the weight for a target and source cluster for users and location respectively. 
A key modeling distinction between $\bs{\alpha}$ and $\bs{\beta}$ is that the latter includes \textit{self}-contribution of the node under consideration and in $\bs{\beta}$ we only aim to capture the contribution by the source-cluster on the particular target-cluster. Therefore $\bs{\beta}$ is calculated similarly as $\bs{\alpha}$ (Equation~\ref{eqn:alpha}) after restricting to only the inter-cluster interactions.

\subsubsection{Overcoming the Curse of Pre-Training}\label{curse}
Transfer learning, by definition, requires the transfer source to be \textit{pre}-trained, i.e. for the information propagation across clusters of user and locations, the set of weights for source should be trained before initiating the transfer. In our setting, we do not extensively train the source parameters separately as these parameters are jointly learned via meta-learning. This could be a severe bottleneck for the cluster-based transfer as it may lead to inaccurate information sharing across clusters as the source parameters are also simultaneously being learned. We reconcile these two by adopting a \textit{checkpoint}-based transfer approach by performing transfer based on the number of epochs for parameter optimization of the model. Specifically, we optimize the region-specific model parameters for $M_t$ epochs with each epoch across the entire source and target data. We \textit{checkpoint} this model state and consider the user-location cluster embeddings to initialize transfer by optimizing the all-region parameters through cluster-alignment loss, $\cm{L}_c$~\cite{what2tran}. Optimizing $\cm{L}_c$ updates the user and POI embedding by backpropagating the difference between similar source and target clusters. We minimize $\cm{L}_c$ via stochastic gradient descent(SGD)~\cite{optimization}. This checkpoint-based optimize-transfer cycle continues for fixed iterations and then weights are later \textit{fine-tuned}~\cite{maml}. The learning process is summarized in Algorithm~\ref{axoalgo}.

%\subsection{Training Methodology}\label{training_method}
%For training, $\bm{U}, \bm{L}, \bs{W}, \bs{b}$ and all MLP parameters are initialized randomly from a normal distribution. Given the complex architecture of \ourm and the heterogeneous-domain nature of data and multiple channels of training(meta- and cluster-based), we adopt a checkpoint-based training procedure as mentioned above. This training procedure ensures that the update gradients, along with region-specific information, also include the cross-domain information~\cite{what2tran}. We also show the advantages of this learning approach in \ref{sensitive}.  We refer to the variant of \ourm that uses the meta-learning procedure along with utilizing the checkpoint-based training to optimize the cluster-alignment loss as \textbf{\ourm-a}.

\subsubsection{Significance of using SSML with Cluster Loss} Here we highlight the importance of the two tasks in SSML -- neighborhood prediction and POI recommendation that are achieved by minimizing the loss functions $\cm{L}^{\cm{U}}_{s}$, $\cm{L}^{\cm{P}}_{s}$, and $\cm{L}_{p}$ respectively for each region. Particularly, the task of neighborhood prediction of each region is a combination of predicting the spatial neighbors of a POI (via $\cm{L}^{\cm{P}}_{s}$) and the social network of a user (via $\cm{L}^{\cm{U}}_{s}$). The former ensures that the embeddings of POIs located within a small geographical area can capture the latent features of the particular that area~\cite{locate, www}. Such a feat is not achievable by minimizing the difference between POIs in a common cluster, as the clusters are determined explicitly from these embeddings whereas the spatial graph is constructed using the distance between POIs, and thus is a better estimate of neighborhoods within a region. Similarly, minimizing $\cm{L}^{\cm{U}}_{s}$ ensures that user embeddings capture the flow of POI-preferences between socially connected users~\cite{ngcf} that cannot be captured via an embedding based clustering. Thus, the task of predicting the neighborhood of a node can lead to better POI recommendations to users closer to a locality.
\section{Experiments}
\label{sec:experiments}
We perform check-in recommendation in the test data to evaluate \ourm across three geo-tagged activity streams from different countries. With our experiments we aim to answer the following research questions:
\begin{itemize}
\item[\textbf{RQ1}] Can \ourm outperform state-of-the art baselines for location recommendation in sparse regions?
\item[\textbf{RQ2}] What are the contributions of different modules in \ourm? 
\item[\textbf{RQ3}] How are the weights in \ourm transferred across regions? 
\item[\textbf{RQ4}] How do hyper-parameters impact the performance of \ourm's family of methods? 
\end{itemize}
All our models are implemented in Tensorflow on a server running Ubuntu 16.04. CPU: Intel(R) Xeon(R) Gold 5118 CPU @ 2.30GHz, RAM: 125GB, and GPU: NVIDIA V100 GPU.
%at \href{https://bit.ly/3ryp5Dk}{https://bit.ly/3ryp5Dk}.

\subsection{Experimental Settings}
\textbf{Dataset Description.}
For our experiments we combine POI data from two popular datasets, Gowalla~\cite{scellato} and Foursquare~\cite{lbsn2vec}, across 12 different regions of varied granularities from United States(US), Japan(JP) and Germany(DE). For each country we construct 4 datasets: one with large check-in data and three with limited data. We adopt a commonly followed data cleaning procedure~\cite{cara, locate} ---for \emph{source datasets}, we filter out locations with less than 10 check-ins, users with less than 10 check-ins and less than 5 connections. For {target datasets}, these thresholds are set at 5, 5 and 2 respectively. Higher criteria is used for source datasets to minimize the effects of noisy data during transfer. 
The statistics of the twelve datasets is given in Table \ref{tab:data} with each acronym denoting the following region: (i) CA: California(US), (ii) WA: Washington(US), (iii) MA: Massachusetts(US), (iv) OH: Ohio(US), (v) TY: Tokyo(JP), (vi) HY: Hyogo(JP), (vii) KY: Kyoto(JP), (viii) AI: Aizu(JP) (ix) NR: North-Rhine Westphalia(DE), (x) BW: Baden-W\"urttemberg(DE), (xi) BE: Berlin(DE) and (xii) BV: Bavaria(DE). We consider CA, TY and NR as the source regions and WA, NY, MA and KY, HY, AI and BV, BW, BE as the corresponding target regions.

\noindent \textbf{Evaluation Protocol:}
For each region we consider first 70\% data, based on the time of check-in, as training, 10\% as validation, and the rest as test data for both Gowalla and Foursquare. For each region, we use the training data to get a list of top-k most probable check-in locations for each user and compare with ground-truth check-ins in test. Note that there is no user or location overlap between source and target regions and for each user we only recommend check-ins located in the specific region. To evaluate the effectiveness of all approaches, we use: \textit{Precision@k} and \textit{NDCG@k}, with $k = 1, 5, 10$ and report confidence intervals based on three independent runs.

\noindent \textbf{Parameter Settings:} For all experiments we adopt a 3 layer architecture for MLPs with dimensions $\varphi_1, \varphi_2 =\{32\rightarrow32 \rightarrow D \}$ and $\varphi_3= \{32\rightarrow32\rightarrow1\}$. Other variations for the MLP had insignificant differences. We keep $M_t = \{4, 6\}, N_u = \{4, 8\}, K = \{20, 50, 100\}, \lambda_p = 0.01$, $\kappa = 50km, D=16$ and batch-size in $\{16, 32\}$. Unless otherwise mentioned, we use these parameters in all our experiments. For the two channels of transfer, meta-learning- and cluster-based, we set $\omega_1, \omega_2, \omega_3 = 0.001$ and learning-rate as $0.01$, as recommended for training a meta-learning algorithm ~\cite{maml}.

\begin{table*}[t!]
\caption{Statistics of datasets used in our experiments. The source region columns are highlighted, followed by target regions. The datasets are further partitioned based on the country of origin (US, Japan and Germany).}
\vspace{-2mm}
\label{tab:data}
\centering
\resizebox{\textwidth}{!}{
\begin{tabular}{l|accc|accc|accc}
\toprule
\textbf{Property} & \textbf{CA} & \textbf{WA} & \textbf{MA} & \textbf{OH} & \textbf{TY} & \textbf{KY} & \textbf{AI} & \textbf{HY} & \textbf{NR} & \textbf{BV} & \textbf{BW} & \textbf{BE}\\
\hline
\#Users ($|\cm{U}|$) & 3518 & 1959 & 1623 & 1322 & 6361 & 1445 & 2059 & 1215 & 1877 & 923 & 682 & 1015\\
\#Locations ($|\cm{P}|$) & 42125 & 16758 & 10585 & 8509 & 11905 & 2055 & 4561 & 2255 & 13049 & 7290 & 8381 & 5493\\
\#User-User Edges ($|\mathcal{E}_u|)$ & 26167 & 5039 & 3702 & 3457 & 32540 & 2973 & 5723 & 2124 &  5663 & 2833 & 2370 & 3122\\
\#Location-Location Edges ($|\mathcal{E}_l|$) & 250360 & 94400 & 46225 & 34182 & 146021 & 15237 & 37470 & 14055 & 66086 & 31823 & 40253 & 29308\\
\#User-Location Edges ($|\mathcal{E}_r|$) & 310350 & 114897 & 52097 & 37431 & 189990 & 17573 & 40503 & 17036 & 75212 & 37673 & 50889 & 31915\\
\bottomrule
\end{tabular}
}
\vspace{-2mm}
\end{table*}

\subsection{Methods} \label{baselines}
We compare \ourm with the state-of-the-art methods based on their architectures below:
\begin{compactenum}[(1)]
\item{\bf Methods based on Random Walks:}
\begin{asparadesc}
%(Grover et. al, 2016)
\item [Node2Vec~\cite{node2vec}] Popular random-walks based embedding approach that uses parameterized breadth- and depth-first search to capture representations.
\item [Lbsn2Vec~\cite{lbsn2vec}] State-of-the-art random-walk based POI recommendation, it uses a random-walk-with-stay scheme to jointly sample user check-ins and social relationships to learn node embeddings.
\end{asparadesc}
\item \textbf{Graph-based POI Recommendation}
\begin{asparadesc}
\item [Reline~\cite{reline}] State-of-the-art multi-graph based POI recommendation algorithm. Traverses across location, user graphs to generate individual embeddings.
\end{asparadesc}

\item \textbf{Methods based on Matrix Factorization:}
\begin{asparadesc}
\item [GMF~\cite{gmf}] Standard matrix factorization which is optimized using a personalized prediction loss for users.
\item [NMF~\cite{ncf}] Collaborative filtering based model that applies MLPs above the concatenation of user and item embeddings to capture their interactions.
\end{asparadesc}

\item \textbf{Methods based on Graph Neural Networks:}
\begin{asparadesc}
\item [NGCF~\cite{ngcf}] State-of-the-art graph neural network recommendation framework that encodes the collaborative signal with connectivity in user-item bipartite graph.
\item [DANSER~\cite{danser}] Uses \textit{dual} graph-attention networks across item and user networks and predicts using a reinforcement policy-based algorithm. 
\end{asparadesc}
\item \textbf{Methods using Transfer Learning:}
\begin{asparadesc}
\item [MDNN~\cite{homanga}] An MLP based meta-learning model that performs \textit{global} as well as \textit{local} updates together.
\item [MCSM~\cite{manasi}] A neural architecture with parameters learned through an optimization-based meta-learning.
\item [MeLU~\cite{melu}] State-of-the-art \textit{meta}-learning based recommendation system. Estimates user preferences in a data-limited query set by using a data-rich support set across the concatenation of user and item representations.
\item [MAMO~\cite{mamo}] Modifies MAML~\cite{maml} to incorporate heterogeneous information network by item content information and memory-based mechanism.
\item [PGN~\cite{pretrain}] A state-of-the-art meta-learning procedure for pre-training neural graph models to better capture the user and item embeddings. Specifically, it includes a three step procedure -- a neighborhood sampler, a GNN-based aggregator, and meta-learning based updates. For our experiments, we apply PGN over graph attention networks~\cite{gat}.
\end{asparadesc}
\end{compactenum}
As mentioned in Section \ref{intro}, other techniques either collectively learn parameters across \textit{common} users in both domains\cite{cmf, compare} or either utilize to meta-path based approach~\cite{mhin}, and thus are not suitable for our setting. Furthermore, to demonstrate the drawbacks of traditional transfer learning, we report results for the following variants of \ourm:
\begin{asparaenum}[(1)]
\item [\textbf{\ours-f:}] We train \ours-basic on the source data and fine-tune the weights for the target data as a standard transfer-learning setting.
\item [\textbf{\ours-m:}] For this variant \ours-basic model is trained on both regions using only the proposed spatio-social meta learning, i.e. without cluster-based optimization.
\end{asparaenum}
The main contribution of the paper that includes both SSML and cluster based optimization is termed as \ourm. 

\noindent \textbf{Baseline Implementations:} Here, we present the implementation details for each of our baselines. Specifically, for region-specific models, we follow a standard practice of optimizing their parameters on the training set of the target region and then predicting for users in the corresponding test set. For MDNN and MCSM, train the parameters on both regions using their standard meta-learning procedure. For MeLU, we modify the training protocol to perform \textit{global-updates} using the user-POI pairs for the target regions and \textit{local-updates} using the source-region parameters. For more details, please refer to Section 3.2 in~\cite{melu}. A similar training procedure is followed for MAMO. Lastly, for PGN we pre-train the model parameters on the source-region checkins and then fine-tune on the target region as per their three step optimizing procedure.

\begin{table*}[t!]
\caption{\label{tab:main} Performance comparison between state-of-the-art baselines, \ourm and its variants (\ours-f and \ours-m). The first column represents the \textit{source} and the corresponding \textit{target} regions. The grouping is done based on baseline details in Section \ref{baselines}. 
%the underlying mechanisms in the order: (i) Random-Walk based; (ii) Graph based; (iii) MF-based; (iv) GCN-based; (v) Transfer-learning based, and (vi) Axolotl variants. 
$\bs{\Delta}$ and $\bs{\Delta_T}$ respectively denote the performance gain over the best performing baseline and the advantage over training only on the target region with no transfer. Numbers with bold font indicate the best performing model. All results marked $\dagger$ are statistically significant (i.e. two-sided Fisher's test with $p \le 0.05$) over the best baseline.}
\vspace{-2mm}
\centering
\resizebox{\textwidth}{!}{
\begin{tabular}{c|c|cc|c|cc|cc|ccccc|aaa}
\toprule
$\bs{\mathcal{D}}^{src} \rightarrow \bs{\mathcal{D}}^{tgt}$ & \textbf{Metric} & \textbf{N2V} & \textbf{L2V} & \textbf{Reline} & \textbf{GMF} & \textbf{NMF} & \textbf{Danser} & \textbf{NGCF} & \textbf{MDNN} & \textbf{MCSM} & \textbf{MeLU} & \textbf{MAMO} & \textbf{PGN} & \textbf{Axo-f} & \textbf{Axo-m} & \textbf{Axolotl}\\ \hline

\multirow{5}{*}{\textbf{CA} $\rightarrow$ \textbf{WA}}
 & Prec@1 & 0.3053 & 0.3493 & 0.3912 & 0.3709 & 0.4306 & 0.4792 & 0.4716 & 0.4639 & 0.4287 & 0.5094 & 0.4783 & 0.5038 & 0.4605 & 0.5529 & \textbf{0.5537}$\dagger$\\
 & Prec@5 & 0.5618 & 0.6042 & 0.6248 & 0.6096 & 0.6630 & 0.7124 & 0.6983 & 0.6733 & 0.6694 & 0.7041 & 0.6974 & 0.7067 & 0.6793 & 0.7714 & \textbf{0.7736}$\dagger$\\ 
 & Prec@10 & 0.5652 & 0.5718 & 0.6198 & 0.5982 & 0.6537 & 0.7012 & 0.6981 & 0.6787 & 0.6436 & 0.7003 & 0.6827 & 0.6941 & 0.6832 & 0.7431 & \textbf{0.7489}$\dagger$\\ 
 \cline{2-17}
 & NDCG@5 & 0.3804 & 0.4533 & 0.5386 & 0.5228 & 0.5688 & 0.5809 & 0.5646 & 0.5743 & 0.5680 & 0.6064 & 0.5814 & 0.5932 & 0.5788 & 0.6473 & \textbf{0.6503}$\dagger$\\ 
 & NDCG@10 & 0.4372 & 0.4976 & 0.6067 & 0.5884 & 0.6461 & 0.6710 & 0.6587 & 0.6511 & 0.6449 & 0.6721 & 0.6687 & 0.6783 & 0.6572 & 0.7199 & \textbf{0.7285}$\dagger$\\ 
\hline

 \multirow{5}{*}{\textbf{CA} $\rightarrow$ \textbf{OH}}
 & Prec@1 & 0.4453 & 0.4787 & 0.5239 & 0.4863 & 0.5055 & 0.5437 & 0.5386 & 0.5035 & 0.5019 & 0.5492 & 0.5238 & 0.5391 & 0.5394 & 0.6083 & \textbf{0.6102}$\dagger$\\
 & Prec@5 & 0.5618 & 0.5694 & 0.6022 & 0.5745 & 0.6152 & 0.6767 & 0.6643 & 0.6283 & 0.6198 & 0.6759 & 0.6693 & 0.6774 & 0.6480 & 0.7118 & \textbf{0.7134}$\dagger$\\ 
 & Prec@10 & 0.5651 & 0.5703 & 0.6173 & 0.5802 & 0.6247 & 0.6759 & 0.6681 & 0.6382 & 0.6233 & 0.6659 & 0.6628 & 0.6648 & 0.6531 & 0.7023 & \textbf{0.7042}$\dagger$\\ 
 \cline{2-17}
 & NDCG@5 & 0.4504 & 0.4576 & 0.4973 & 0.4581 & 0.4850 & 0.5017 & 0.5132 & 0.5271 & 0.5136 & 0.5498 & 0.5231 & 0.5370 & 0.5344 & 0.6088 & \textbf{0.6110}$\dagger$\\ 
 & NDCG@10 & 0.5070 & 0.50934 & 0.5658 & 0.5781 & 0.5840 & 0.6129 & 0.6115 & 0.6049 & 0.5934 & 0.6180 & 0.5890 & 0.6134 & 0.6222 & 0.6654 & \textbf{0.6690}$\dagger$\\
\hline

 \multirow{5}{*}{\textbf{CA} $\rightarrow$ \textbf{MA}}
 & Prec@1 & 0.3964 & 0.3900 & 0.4239 & 0.4092 & 0.4476 & 0.4886 & 0.4761 & 0.4462 & 0.4581 & 0.4889 & 0.4731 & 0.4807 & 0.4539 & \textbf{0.5813}$\dagger$ & 0.5807\\
 & Prec@5 & 0.5618 & 0.5737 & 0.6282 & 0.5816 & 0.6273 & 0.6873 & 0.6791 & 0.6432 & 0.6400 & 0.6808 & 0.6814 & 0.6973 & 0.6400 & 0.7349 & \textbf{0.7373}$\dagger$\\ 
 & Prec@10 & 0.5651 & 0.5899 & 0.6215 & 0.6007 & 0.6266 & 0.6798 & 0.6620 & 0.6419 & 0.6302 & 0.6649 & 0.6587 & 0.6798 & 0.6545 & 0.7316 & \textbf{0.7317}$\dagger$\\
 \cline{2-17}
 & NDCG@5 & 0.4206 & 0.4491 & 0.5243 & 0.4860 & 0.5034 & 0.5614 & 0.5587 & 0.5219 & 0.5187 & 0.5759 & 0.5602 & 0.5683 & 0.5342 & 0.6392 & \textbf{0.6404}$\dagger$\\
 & NDCG@10 & 0.4785 & 0.5084 & 0.5528 & 0.5382 & 0.5476 & 0.6186 & 0.6037 & 0.5764 & 0.5622 & 0.6294 & 0.6140 & 0.6381 & 0.5785 & 0.6930 & \textbf{0.6941}$\dagger$\\
\hline
 
 \multirow{5}{*}{\textbf{TY} $\rightarrow$ \textbf{AI}}
 & Prec@1 & 0.4703 & 0.5407 & 0.6130 & 0.5811 & 0.6026 & 0.5930 & 0.6200 & 0.5960 & 0.6018 & 0.6462 & 0.6137 & 0.6390 & 0.6625 & 0.7190 & \textbf{0.7218}$\dagger$\\
 & Prec@5 & 0.5411 & 0.6008 & 0.6119 & 0.6044 & 0.6289 & 0.6519 & 0.6579 & 0.5996 & 0.6433 & 0.6610 & 0.6413 & 0.6583 & 0.6389 & 0.7216 & \textbf{0.7231}$\dagger$\\
 & Prec@10 & 0.5926 & 0.6122 & 0.6208 & 0.6135 & 0.6333 & 0.6540 & 0.6569 & 0.6152 & 0.6172 & 0.6430 & 0.6217 & 0.6535 & 0.6522 & \textbf{0.7090}$\dagger$ & 0.7017\\
 \cline{2-17}
 & NDCG@5 & 0.4489 & 0.5013 & 0.5139 & 0.5032 & 0.5390 & 0.5346 & 0.5958 & 0.5304 & 0.5799 & 0.6174 & 0.5689 & 0.6003 & 0.5850 & 0.6581 & \textbf{0.6631}$\dagger$\\
 & NDCG@10 & 0.5562 & 0.5796 & 0.5927 & 0.5818 & 0.6128 & 0.6145 & 0.6485 & 0.5802 & 0.6004 & 0.6298 & 0.5910 & 0.6344 & 0.6299 & \textbf{0.7082}$\dagger$ & 0.7052\\  
\hline

 \multirow{5}{*}{\textbf{TY} $\rightarrow$ \textbf{KY}}
 & Prec@1 & 0.4317 & 0.4812 & 0.5226 & 0.4926 & 0.5837 & 0.5719 & 0.5914 & 0.5518 & 0.5867 & 0.6173 & 0.5874 & 0.6074 & 0.6209 & 0.6847 & \textbf{0.6910}$\dagger$\\
 & Prec@5 & 0.5811 & 0.6002 & 0.6044 & 0.5869 & 0.6372 & 0.6579 & 0.6567 & 0.6213 & 0.6014 & 0.6521 & 0.6041 & 0.6794 & 0.6348 & 0.7011 & \textbf{0.7018}$\dagger$\\
 & Prec@10 & 0.5926 & 0.6098 & 0.6123 & 0.5995 & 0.6243 & 0.6555 & 0.6569 & 0.6135 & 0.6005 & 0.6695 & 0.6117 & 0.6741 & 0.6287 & 0.7019 & \textbf{0.7093}$\dagger$\\
 \cline{2-17}
 & NDCG@5 & 0.4789 & 0.5024 & 0.5016 & 0.5120 & 0.5878 & 0.5758 & 0.6291 & 0.6024 & 0.5818 & 0.6892 & 0.6499 & 0.6742 & 0.6458 & 0.7390 & \textbf{0.7454}$\dagger$\\
 & NDCG@10 & 0.5562 & 0.5752 & 0.5807 & 0.5621 & 0.6271 & 0.6485 & 0.6500 & 0.6318 & 0.6044 & 0.7192 & 0.6786 & 0.6983 & 0.6857 & 0.7720 & \textbf{0.7737}$\dagger$\\
\hline

 \multirow{5}{*}{\textbf{TY} $\rightarrow$ \textbf{HY}}
 & Prec@1 & 0.5208 & 0.5719 & 0.5832 & 0.5582 & 0.6030 & 0.6139 & 0.6083 & 0.5990 & 0.5752 & 0.6146 & 0.5824 & 0.6108 & 0.5854 & \textbf{0.6793}$\dagger$ & 0.6781\\
 & Prec@5 & 0.5411 & 0.5996 & 0.5938 & 0.6053 & 0.6018 & 0.6527 & 0.6719 & 0.6419 & 0.6322 & 0.6685 & 0.6459 & 0.6814 & 0.6484 & \textbf{0.7172}$\dagger$ & 0.7134\\
 & Prec@10 & 0.5726 & 0.6072 & 0.6059 & 0.6205 & 0.6138 & 0.6541 & 0.6569 & 0.6348 & 0.6206 & 0.6406 & 0.6283 & 0.6746 & 0.6268 & 0.6994 & \textbf{0.7019}$\dagger$\\ 
 \cline{2-17}
 & NDCG@5 & 0.4891 & 0.4959 & 0.4888 & 0.5316 & 0.6183 & 0.5724 & 0.6195 & 0.6092 & 0.6151 & 0.6480 & 0.6204 & 0.6359 & 0.5968 & 0.7043 & \textbf{0.7068}$\dagger$\\
 & NDCG@10 & 0.5262 & 0.5736 & 0.5681 & 0.6047 & 0.6256 & 0.6461 & 0.6485 & 0.6380 & 0.6353 & 0.6793 & 0.6433 & 0.6684 & 0.6225 & 0.7298 & \textbf{0.7352}$\dagger$\\ 
\hline

\multirow{5}{*}{\textbf{NR} $\rightarrow$ \textbf{BE}}
 & Prec@1 & 0.4624 & 0.4721 & 0.4689 & 0.4627 & 0.4830 & 0.5211 & 0.5354 & 0.5277 & 0.5101 & 0.5471 & 0.5371 & 0.5439 & 0.5392& 0.5972 & \textbf{0.5981}$\dagger$\\
 & Prec@5 & 0.4680 & 0.4981 & 0.5783 & 0.5436 & 0.5836 & 0.6430 & 0.6214 & 0.6023 & 0.6148 & 0.6471 & 0.6252 & 0.6572 & 0.6392 & \textbf{0.6900}$\dagger$ & 0.6889\\ 
 & Prec@10 & 0.5254 & 0.5634 & 0.6092 & 0.5999 & 0.6124 & 0.6444 & 0.6250 & 0.6162 & 0.6029 & 0.6391 & 0.6281 & 0.6492 & 0.6215 & 0.6742 & \textbf{0.6749}$\dagger$\\ 
 \cline{2-17}
 & NDCG@5 & 0.4998 & 0.5077 & 0.5315 & 0.5189 & 0.5416 & 0.5920 & 0.6018 & 0.6297 & 0.6291 & 0.6603 & 0.6198 & 0.6569 & 0.6631 & 0.7370 & \textbf{0.7376}$\dagger$\\ 
 & NDCG@10 & 0.5532 & 0.5674 & 0.6285 & 0.5779 & 0.6248 & 0.6826 & 0.6604 & 0.6633 & 0.6516 & 0.6889 & 0.6749 & 0.6842 & 0.6811 & 0.7734 & \textbf{0.7741}$\dagger$\\ 
\hline

 \multirow{5}{*}{\textbf{NR} $\rightarrow$ \textbf{BV}}
 & Prec@1 & 0.3284 & 0.3794 & 0.4024 & 0.3878 & 0.4115 & 0.4588 & 0.4308 & 0.4465 & 0.4209 & 0.4896 & 0.4482 & 0.5163 & 0.4745 & 0.5562 & \textbf{0.5593}$\dagger$\\
 & Prec@5 & 0.4380 & 0.4559 & 0.5005 & 0.4843 & 0.4917 & 0.5649 & 0.5318 & 0.5250 & 0.5075 & 0.5657 & 0.5318 & 0.5739 & 0.5736 & 0.6296 & \textbf{0.6323}$\dagger$\\ 
 & Prec@10 & 0.4860 & 0.5062 & 0.5324 & 0.5150 & 0.5456 & 0.5882 & 0.5675 & 0.5744 & 0.5700 & 0.5809 & 0.5624 & 0.6044 & 0.5648 & \textbf{0.6357}$\dagger$ & 0.6355\\ 
 \cline{2-17}
 & NDCG@5 & 0.4378 & 0.4543 & 0.4713 & 0.4571 & 0.4868 & 0.5239 & 0.5243 & 0.5303 & 0.5271 & 0.5694 & 0.5473 & 0.5789 & 0.5700 & 0.6183 & \textbf{0.6226}$\dagger$\\ 
 & NDCG@10 & 0.4772 & 0.4886 & 0.5768 & 0.5131 & 0.5847 & 0.6043 & 0.5982 & 0.5906 & 0.6019 & 0.6232 & 0.5980 & 0.6267 & 0.6191 & 0.6708 & \textbf{0.6739}$\dagger$\\  
\hline
 
 \multirow{5}{*}{\textbf{NR} $\rightarrow$ \textbf{BW}}
 & Prec@1 & 0.3584 & 0.3858 & 0.4172 & 0.3966 & 0.4263 & 0.4767 & 0.4594 & 0.4520 & 0.4363 & 0.4773 & 0.4587 & 0.4758 & 0.4689 & \textbf{0.5066}$\dagger$ & 0.5062\\
 & Prec@5 & 0.4861 & 0.5197 & 0.5416 & 0.5282 & 0.5579 & 0.5670 & 0.5683 & 0.5564 & 0.5423 & 0.5849 & 0.5670 & 0.5913 & 0.5740 & 0.6198 & \textbf{0.6201}$\dagger$\\
 & Prec@10 & 0.4860 & 0.5487 & 0.5693 & 0.5354 & 0.5293 & 0.5878 & 0.5849 & 0.5694 & 0.5459 & 0.5805 & 0.5718 & 0.5866 & 0.5712 & \textbf{0.6306}$\dagger$ & 0.6297\\ 
 \cline{2-17}
 & NDCG@5 & 0.4102 & 0.4972 & 0.5122 & 0.4898 & 0.5167 & 0.5750 & 0.5731 & 0.5460 & 0.5281 & 0.5623 & 0.5372 & 0.5683 & 0.5563 & 0.6048 & \textbf{0.6075}$\dagger$\\ 
 & NDCG@10 & 0.4772 & 0.5532 & 0.5783 & 0.5552 & 0.5862 & 0.6217 & 0.6101 & 0.5962 & 0.5931 & 0.6183 & 0.5849 & 0.6283 & 0.6000 & 0.6598 & \textbf{0.6614}$\dagger$\\
\bottomrule
\end{tabular}
}
\vspace{-4mm}
\end{table*}

\subsection{Performance Comparison (RQ1)}
We report on the performance of location recommendation of different methods across all our target datasets in Table~\ref{tab:main}. From these results we make the following key observations:
\begin{asparaitem}[$\bullet$]
\item \ourm, and its variant \ours-m that employ meta-learning, consistently yield the best performance on all the datasets. In particular, the complete \ourm improves over the strongest baselines by 5-18\% across the metrics. These results further signify the importance of a meta-learning based procedure using external data to design solution for limited-data regions.

\item \ours-f does not perform on par with other approaches. This observation further cements the advantage of a joint-learning over traditional fine-tuning. The performance gain by \ourm over \ours-m highlights the importance of minimizing the divergence between the embeddings across the two regions.

\item Among meta-learning-based models, we note that MeLU~\cite{melu} and PGN~\cite{pretrain} perform better than other baseline models, however, are easily outperformed by \ourm. We also note that though \ourm and PGN are graph-based meta-learning models, the performance difference can be attributed to the ability of \ourm to include node-features. Specifically, PGN only leverages the graph structure and cannot thus incorporate any heterogeneous auxiliary information about the entities such as POI category and distances, whereas \ourm captures all features of a spatial network. 

\item The characteristic of MeLU~\cite{melu} to include samples from data-rich network into its meta-learning based procedure leads to significant improvements over other baselines even with its MLP based architecture. These improvements are more noteworthy for smaller datasets like Bavaria (BV). However, with inclusion of graph attention networks, \ourm captures the complex user-location dynamics better than MeLU.

\item Danser~\cite{danser} and NGCF~\cite{ngcf} perform comparable to meta-learning based baselines in some datasets --e.g., MA and WA. This is due to a sufficiently moderate dataset-size to fuel their \textit{dual} graph neural networks. 
%The \textit{dual}-architecture easily traverses the entire heterogeneous graph, which in turn increases the \textit{effective} nodes count, however, we later show that this can also be improved through \ourm. 
Danser~\cite{danser} particularly incorporates a reinforcement learning-based policy optimization which particularly leads to better modeling power albeit at the cost of more computation. However, for extremely limited-data regions and Precision@1 predictions, the input-data size alone is not sufficient to accurately train all parameters.

\item Despite Reline~\cite{reline} being the state-of-the-art multi-graph based model for location recommendation, other methods that incorporate complex structures using dual-GCNs or meta-learning are able to easily outperform it, even under sparse data conditions.
\end{asparaitem}

\noindent To sum up, our empirical analysis suggests the following: \begin{inparaenum}[(i)] \item the state-of-the-art models, including fine-tuning based transfer approaches are not suitable for location recommendation in a limited-data region, \item \ourm is a powerful recommender system not only for mobility networks with limited-data, but also in general, and \item for data-scarce regions, forcing embeddings to adapt as per their clusters that share similar preferences across different regions has significant performance gains. \end{inparaenum}

\subsubsection{Impact of Region Data Size on \ourm} \label{analysis}
\begin{figure}[t!]
\centering
\begin{subfigure}{0.3\columnwidth}
  \centering
  \includegraphics[width=\linewidth, height=3cm]{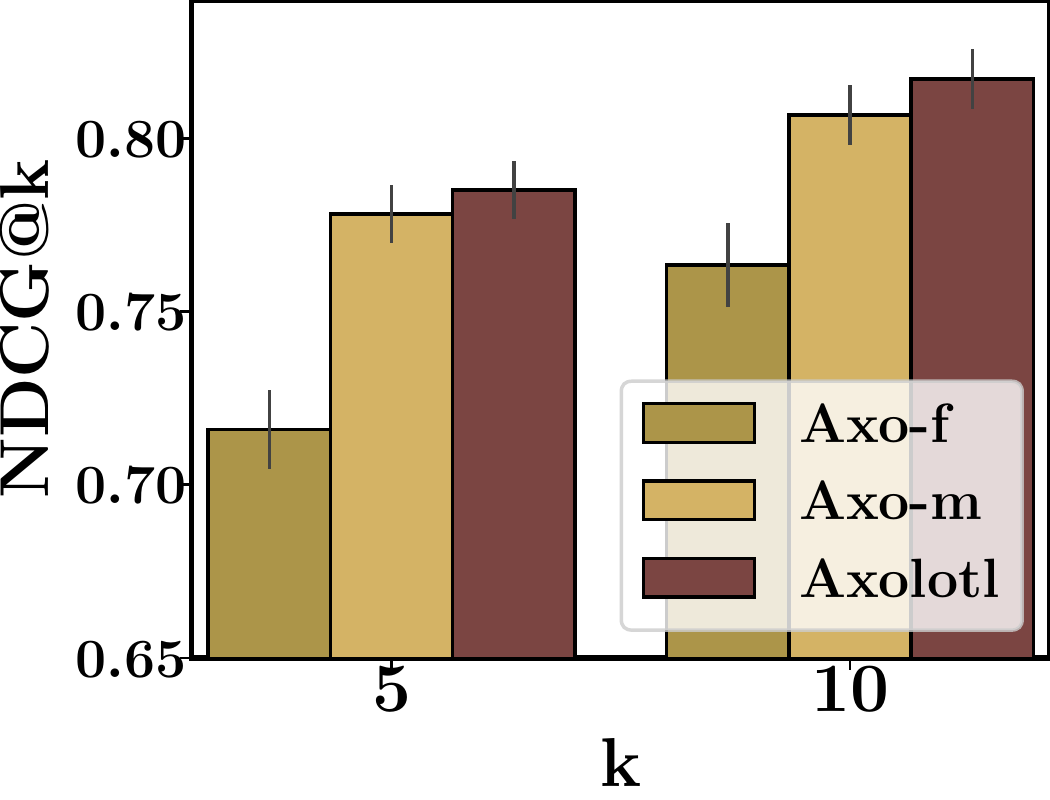}
  \caption{CA $\rightarrow$ WA (40\%)}
  \label{fig:wa_40}
\end{subfigure}
\hfill
\begin{subfigure}{0.3\columnwidth}
  \centering
  \includegraphics[width=\linewidth, height=3cm]{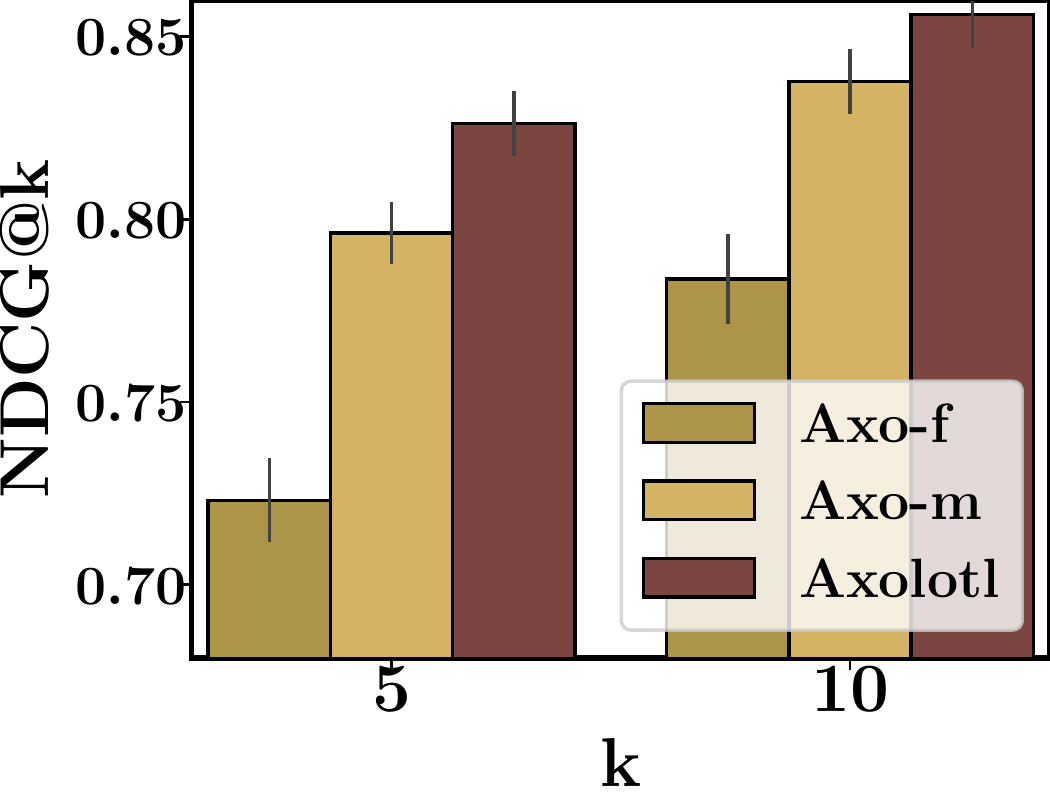}
  \caption{TY $\rightarrow$ AI (40\%)}
  \label{fig:wa_60}
\end{subfigure}
\hfill
\begin{subfigure}{0.3\columnwidth}
  \centering
  \includegraphics[width=\linewidth, height=3cm]{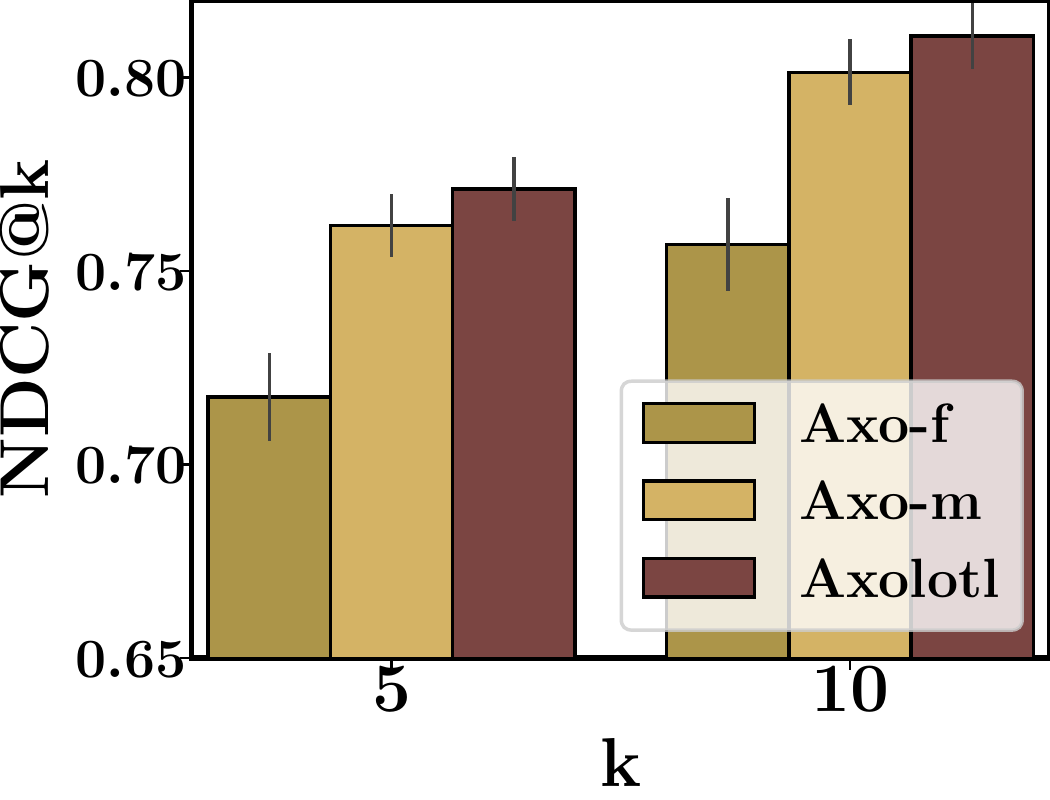}
  \caption{NR $\rightarrow$ BE (40\%)}
  \label{fig:ny_40}
\end{subfigure}
\begin{subfigure}{0.3\columnwidth}
  \centering
  \includegraphics[width=\linewidth, height=3cm]{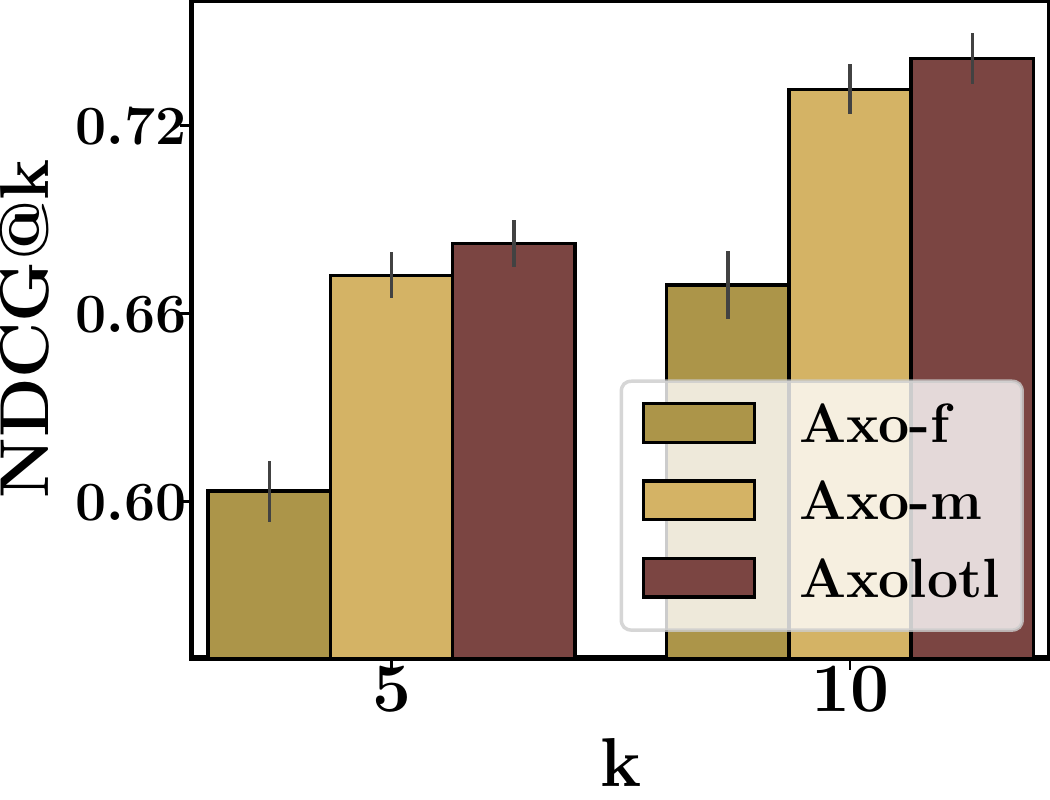}
  \caption{CA $\rightarrow$ WA (60\%)}
  \label{fig:ny_60}
\end{subfigure}
\hfill
\begin{subfigure}{0.3\columnwidth}
  \centering
  \includegraphics[width=\linewidth, height=3cm]{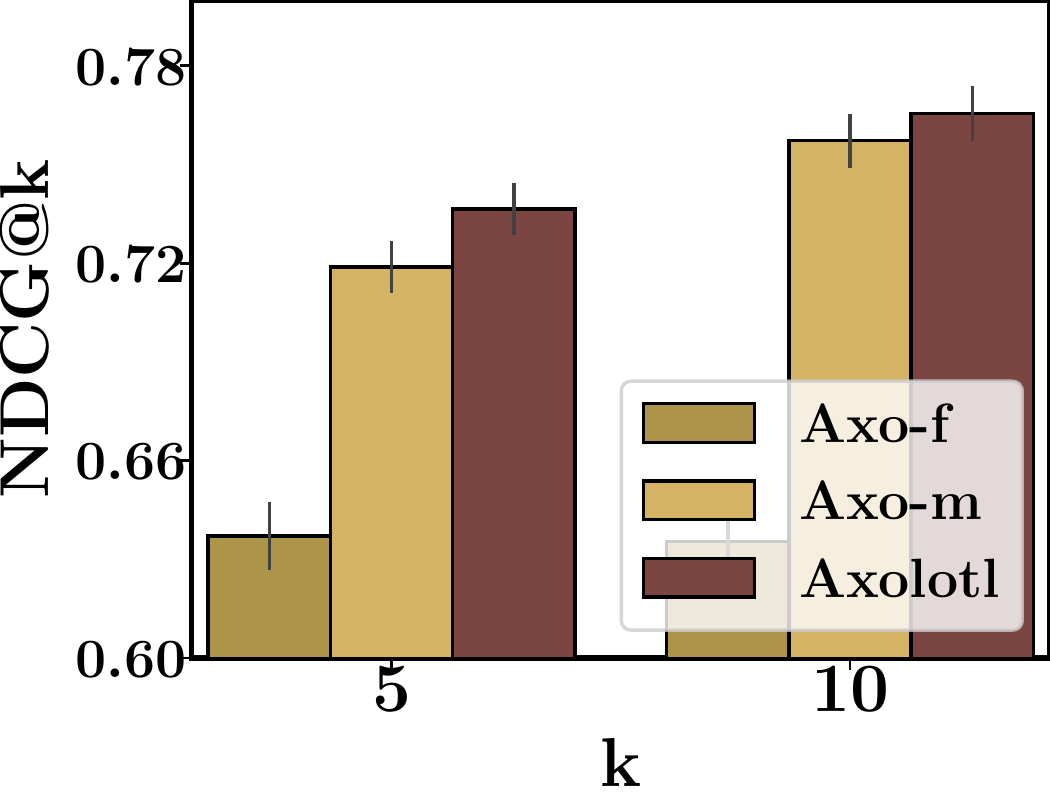}
  \caption{TY $\rightarrow$ AI (40\%)}
  \label{fig:ma_40}
\end{subfigure}
\hfill
\begin{subfigure}{0.3\columnwidth}
  \centering
  \includegraphics[width=\linewidth, height=3cm]{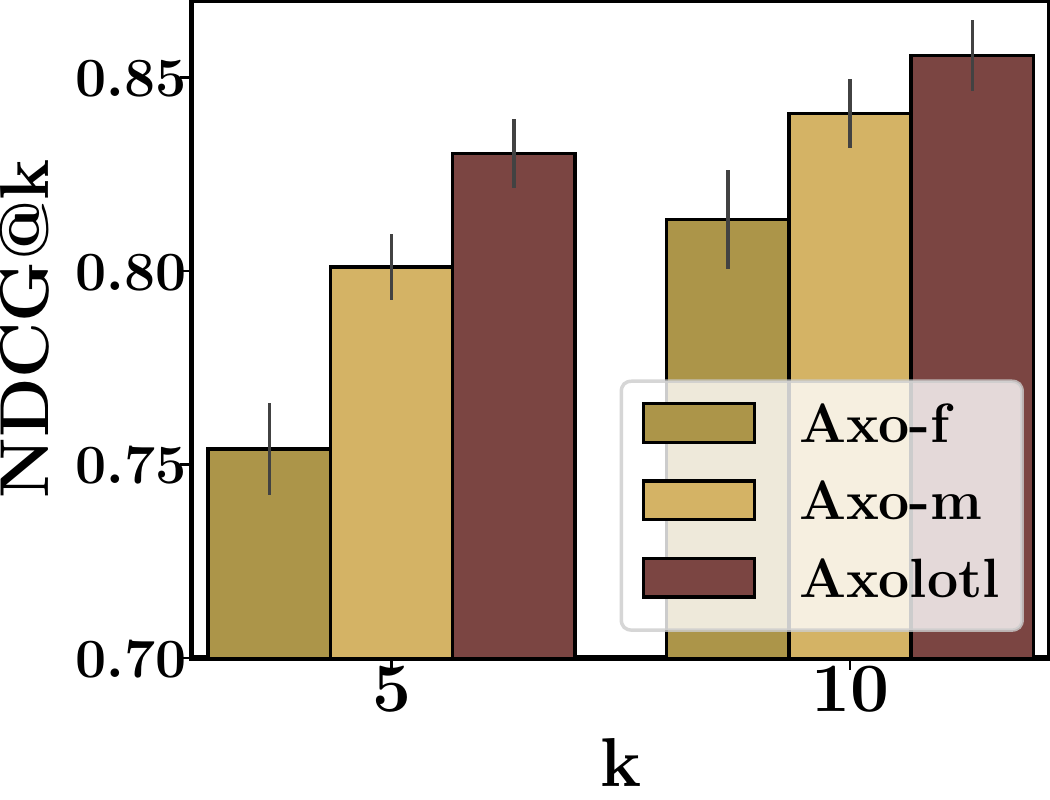}
  \caption{NR $\rightarrow$ BE (60\%)}
  \label{fig:ma_60}
\end{subfigure}
\vspace{-3mm}
\caption{Analysis of \ourm  and its variants across different sizes of training data for both $\ds{D}^{src}$ and $\ds{D}^{tgt}$. Since the results are \textit{transductive}, i.e. we only predict for users and locations present during training, they are specific to the subset of train-data.}
\label{fig:analysis}
\vspace{-3mm}
\end{figure}

Next we turn our attention to evaluating \ourm and its variants under different training data-sizes. Specifically, for each variant, we train with a pre-defined subset of train data, i.e. either 40\% or 60\%, for source as well as target regions and predict on the respective test-set for the target region. We evaluate on the basis of \textit{NDCG@5} and \textit{NDCG@10}. Note that in this section, our prediction setting is \textit{transductive}, i.e. we only predict for those users and locations present in the train data. 
%Therefore, there will be a different user-location set fed into \ourm for training across different training data-sizes. 
Hence the results are specific to the subset of train data used. From the results in Figure \ref{fig:analysis}, we conclude that \ourm and \ours-m consistently outperform \ours-f which further substantiate that \textit{vanilla} transfer is insufficient and minimizing embedding divergence in \ourm leads to better performance in data-scarce regions.

\subsection{Ablation Study (RQ2)}\label{ablation}
\begin{figure}[b]
\centering
\begin{subfigure}{0.3\columnwidth}
  \centering
  \includegraphics[width=\linewidth, height=3cm]{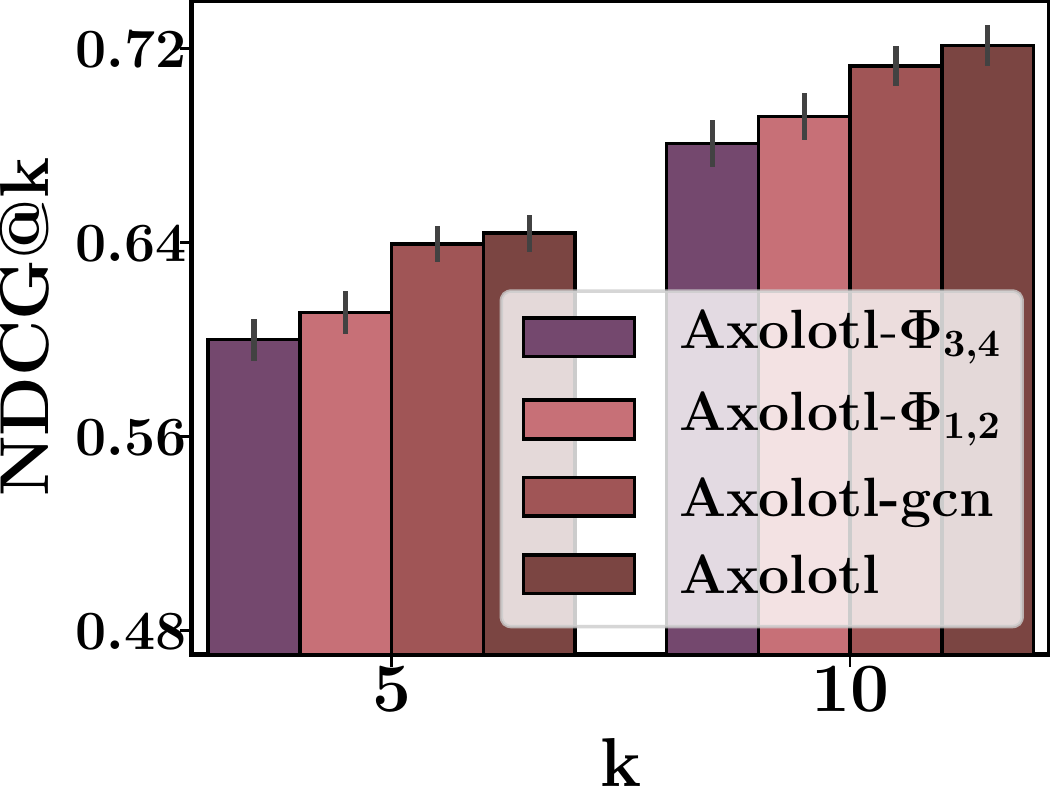}
  \caption{Washington (US)}
\end{subfigure}
\hfill
\begin{subfigure}{0.3\columnwidth}
  \centering
  \includegraphics[width=\linewidth, height=3cm]{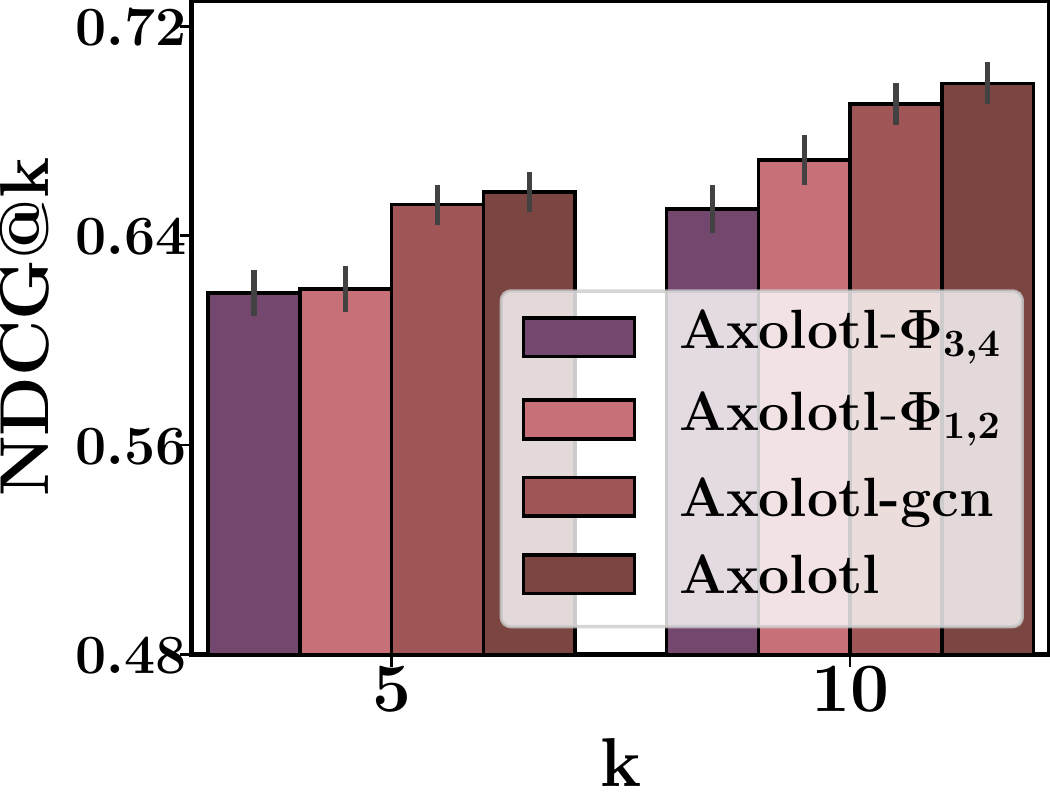}
  \caption{Aizu (JP)}
\end{subfigure}
\hfill
\begin{subfigure}{0.3\columnwidth}
  \centering
  \includegraphics[width=\linewidth, height=3cm]{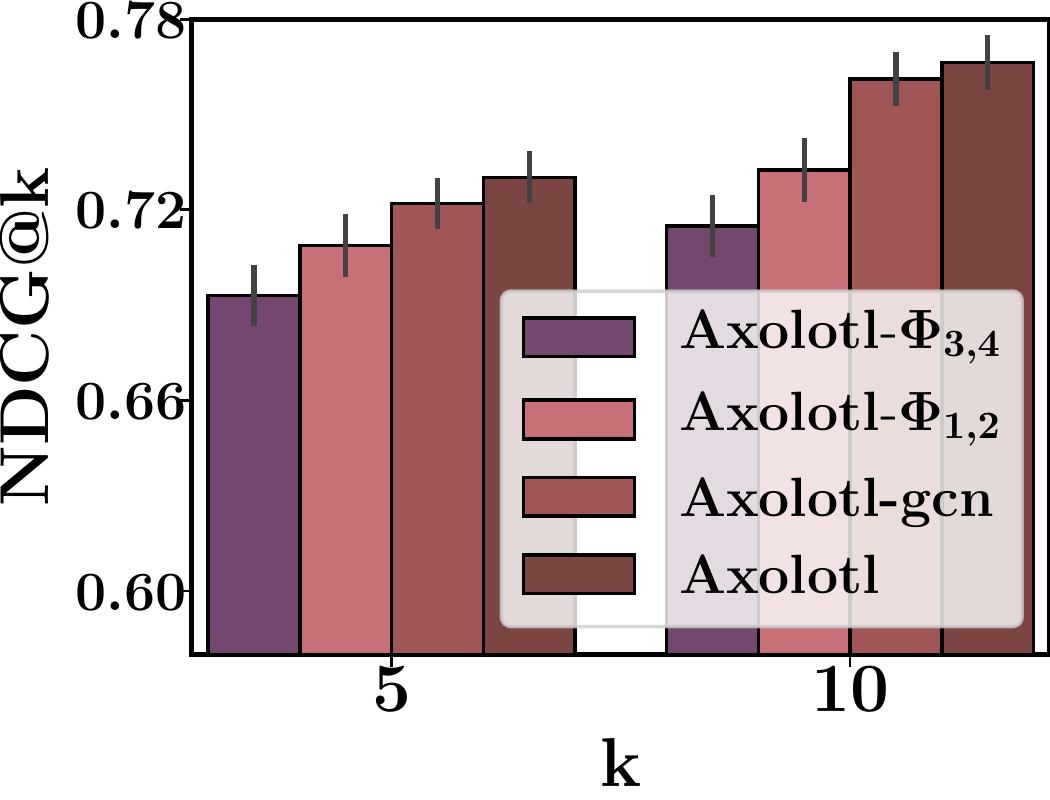}
  \caption{Berlin (DE)}
\end{subfigure}
\vspace{-3mm}
\caption{Ablation Study across different graph architectures possible in \ourm along with a GCN variant.}
\label{fig:ablation}
\vspace{-2mm}
\end{figure}

We conduct an ablation study for two key contributions in \ourm: user-GATs ($\Phi_1$ and $\Phi_2$) and location-GATs ($\Phi_3$ and $\Phi_4$). For estimating the contribution of user-GATs, we use the meta-learning and alignment loss based training only for $\Phi_1$ and $\Phi_2$. We denote this variant as \ourm-$\Phi_{1,2}$. Similarly for location-GATs we use \ourm-$\Phi_{3,4}$. We also include a GCN~\cite{gcn} based implementation of \ourm denoted as \ourm-gcn. From the results in Figure \ref{fig:ablation}, we observe that \ourm with joint training of user and location GATs has better prediction performance than \ourm-$\Phi_{1,2}$ and \ourm-$\Phi_{3,4}$. Interestingly, transferring across users leads to better prediction performance than transferring across locations. This could be attributed to a larger difference in the number of locations between source and target regions in comparison to the number of users. \ourm-gcn has significant performance improvements over the preceding approaches with \ourm further having modest improvements due to weighted neighborhood aggregation.

\begin{figure}[t]
\centering
\begin{subfigure}{0.3\columnwidth}
  \centering
  \includegraphics[width=\linewidth, height=3cm]{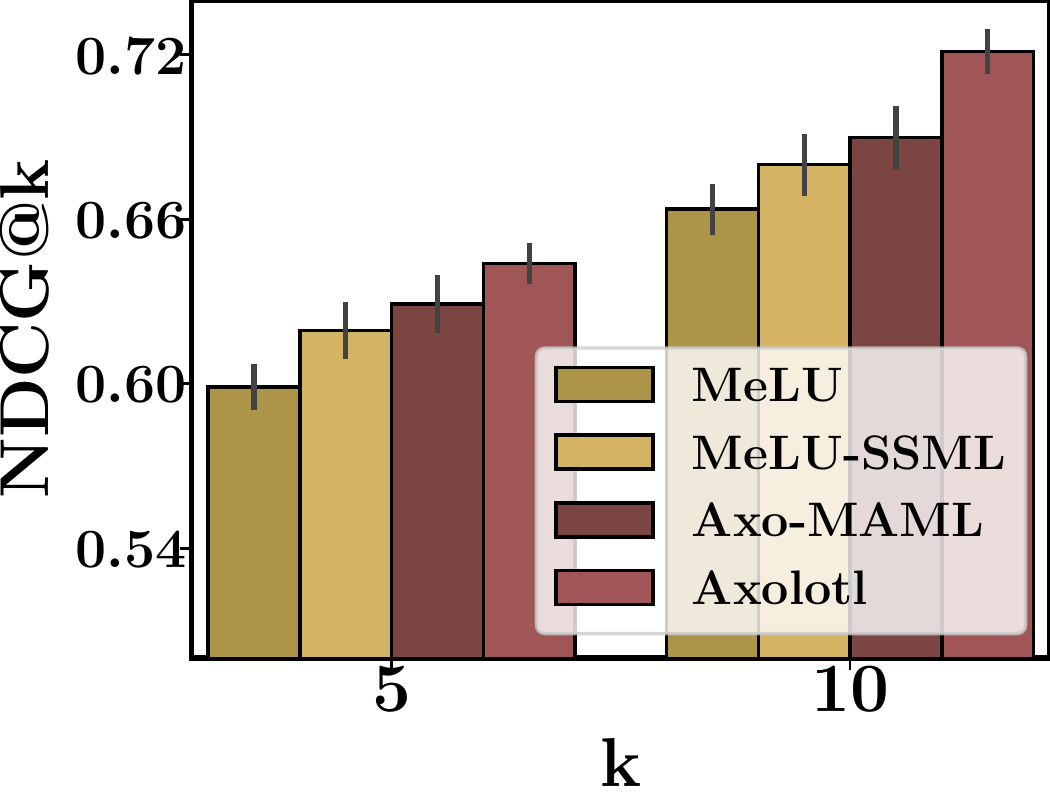}
  \caption{Washington (US)}
\end{subfigure}
\hfill
\begin{subfigure}{0.3\columnwidth}
  \centering
  \includegraphics[width=\linewidth, height=3cm]{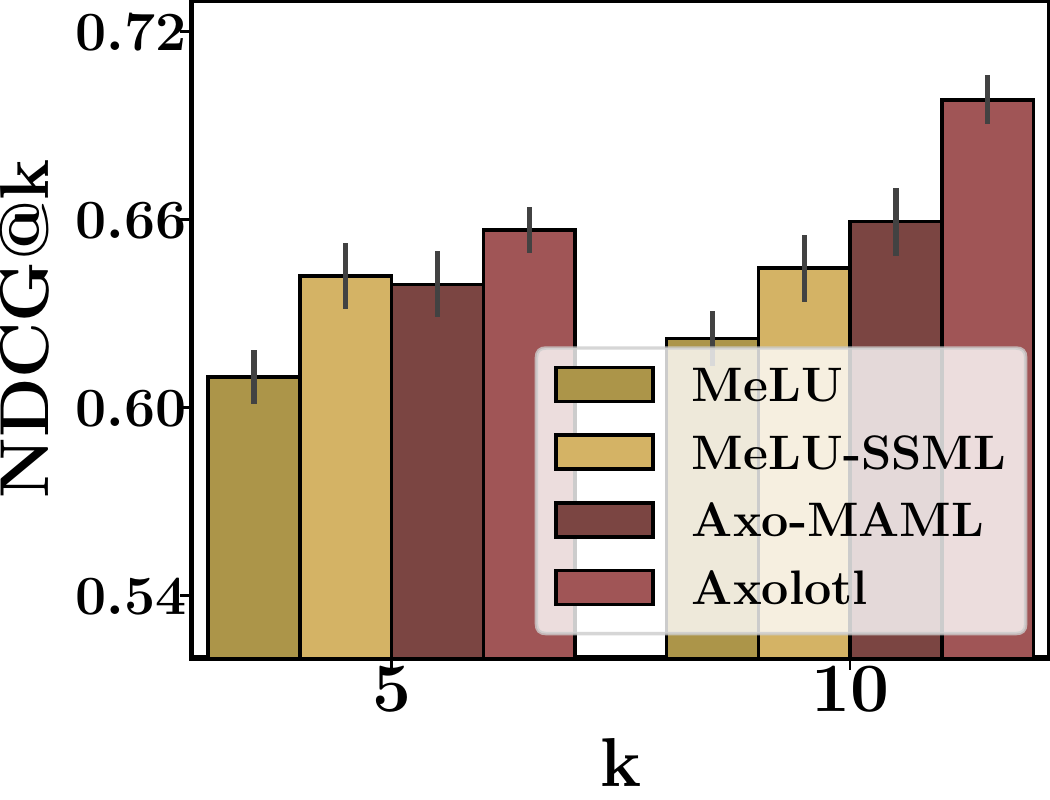}
  \caption{Aizu (JP)}
\end{subfigure}
\hfill
\begin{subfigure}{0.3\columnwidth}
  \centering
  \includegraphics[width=\linewidth, height=3cm]{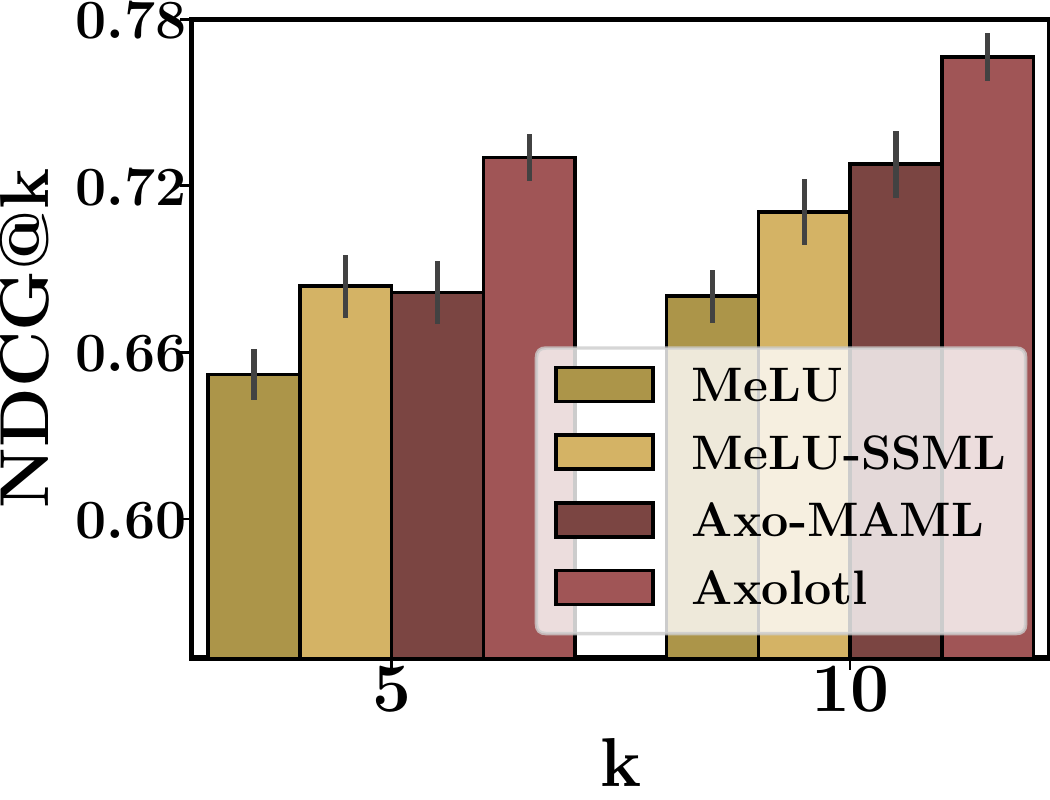}
  \caption{Berlin (DE)}
\end{subfigure}
\vspace{-3mm}
\caption{Contribution of novel spatio-social meta-learning in \ourm and the corresponding gains over MeLU.}
\label{fig:pnp}
\end{figure}

\noindent \textbf{Contribution of SSML:} To further assert the importance of our SSML (Section~\ref{meta_part}), we compare the state-of-the-art MAML-based model, viz., MeLU, with our proposed learning procedure, and also include results after training \ours-basic with MAML. From the results given in Figure \ref{fig:pnp}, we note that MeLU, when trained with SSML, easily outperforms its MAML based counterpart. It not only demonstrates the effectiveness of the proposed learning method, but also its versatility to be incorporated with other baselines. This claim is substantiated further by the poorer performance of \ours-basic with MAML over the complete \ourm model. 

\subsection{Transfer of Weights across Regions (RQ3)}\label{tran_weight}
\begin{figure}[t]
\centering
\begin{subfigure}{0.3\columnwidth}
  \centering
  \includegraphics[width=\linewidth, height=3cm]{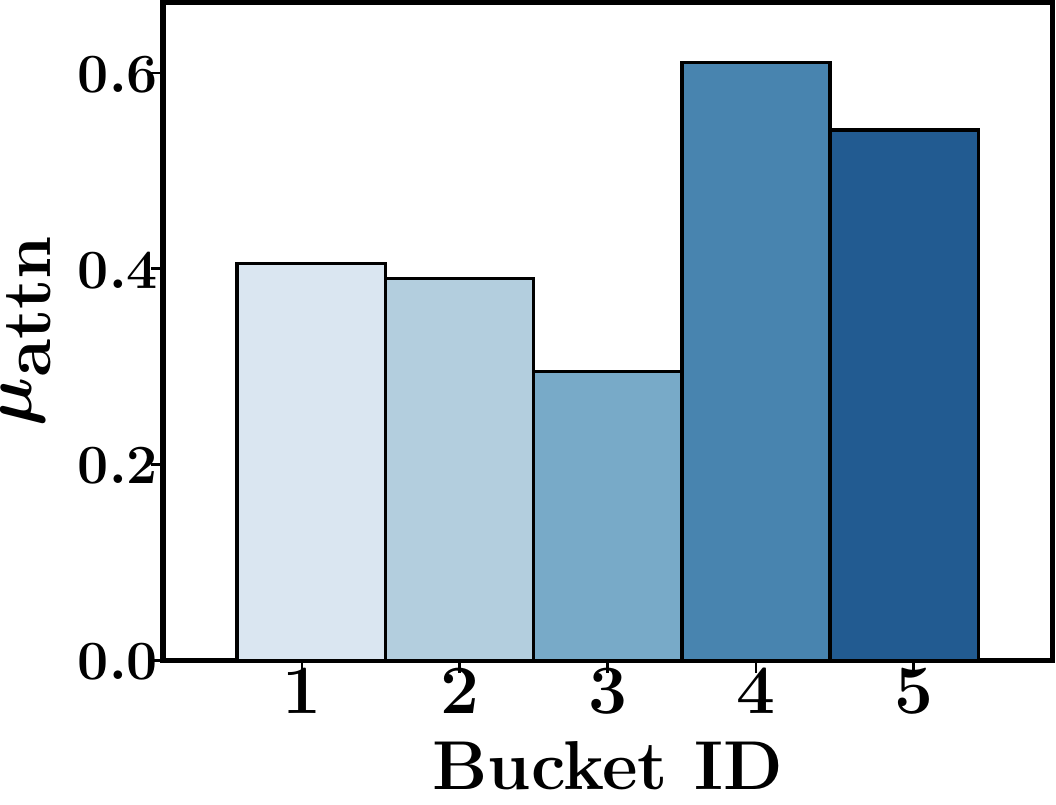}
  \caption{Washington (US)}
\end{subfigure}
\hfill
\begin{subfigure}{0.3\columnwidth}
  \centering
  \includegraphics[width=\linewidth, height=3cm]{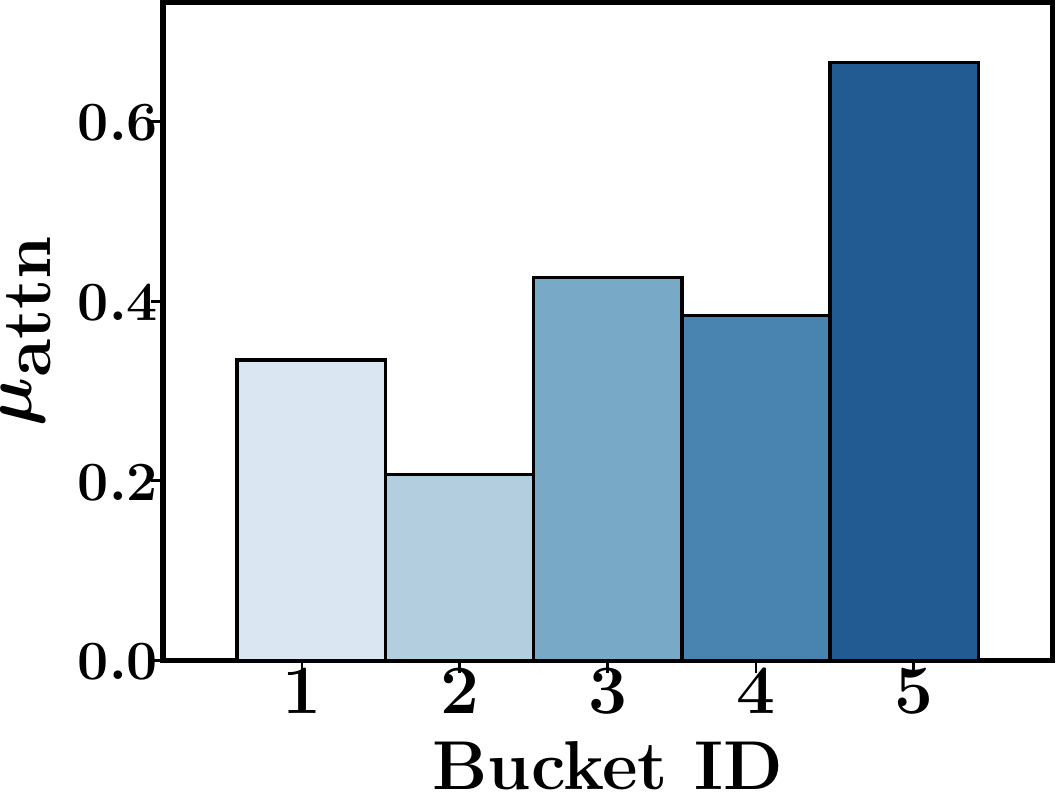}
  \caption{Aizu (JP)}
\end{subfigure}
\hfill
\begin{subfigure}{0.3\columnwidth}
  \centering
  \includegraphics[width=\linewidth, height=3cm]{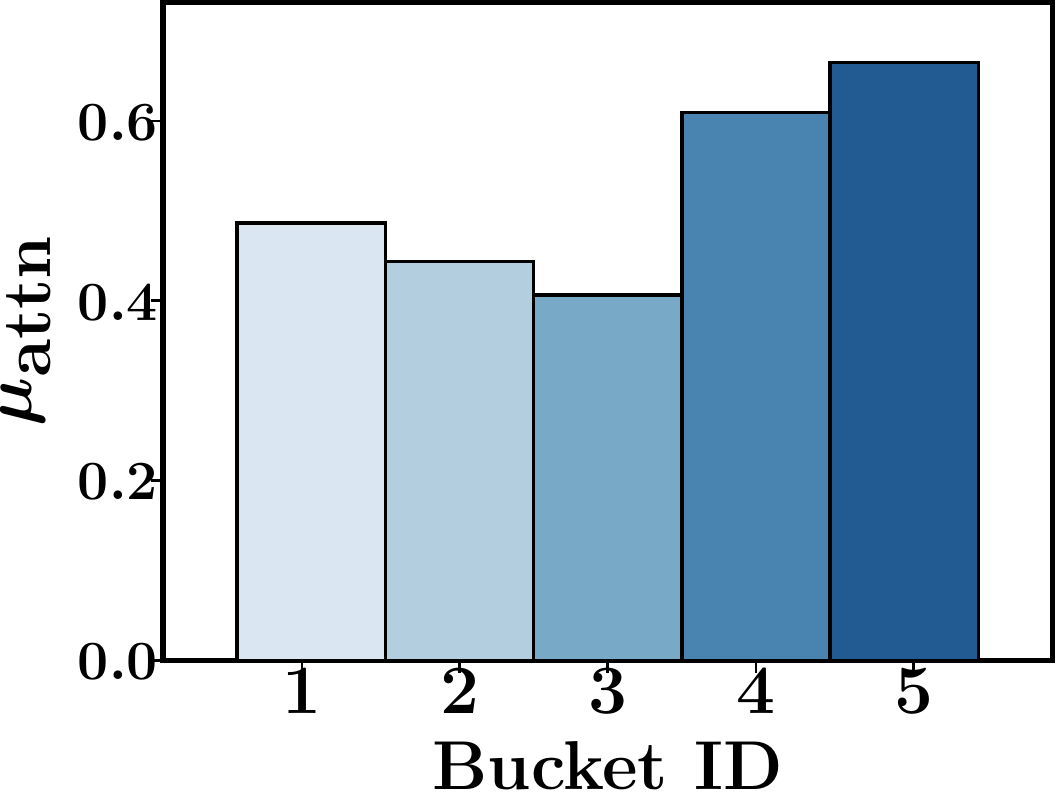}
  \caption{Berlin (DE)}
\end{subfigure}
\vspace{-3mm}
\caption{Bucket-wise Mean Attention Transfer Weights between Source and Target regions.}
\label{fig:tran_weights}
\vspace{-2mm}
\end{figure}

Another important contribution we make in this paper is the cross-region transfer via cluster-based alignment loss. We show that \ourm encapsulates the cluster-wise analogy by plotting the attention-weights corresponding to the similarity between location clusters ($\bm{L}^{tgt}_c$ and $\bm{L}^{src}_c$). We quantify the similarity using \textit{Damerau-Levenshtein} distance~\cite{dldistance} across category distribution for all clusters in source and target regions. Later, we group them into \textit{five} equal buckets as per their similarity score, i.e. bucket-5 will have clusters with higher similarity as compared to other buckets. Figure \ref{fig:tran_weights} shows the mean attention value across each bucket for all datasets. We observe that \ourm is able to capture the increase in inter-cluster similarity by increasing its attention weights. This feature is more prominent for Aizu and comparable for Berlin.
%For Berlin, some buckets with lower similarity have higher attention weight which could be due to other cluster-specific aspects such as POI density, which are beyond category information.\sbcomment{the last statement about Berlin is a bit vague.}

\begin{figure}[b]
\centering
\begin{subfigure}{0.3\columnwidth}
  \centering
  \includegraphics[width=\linewidth, height=3cm]{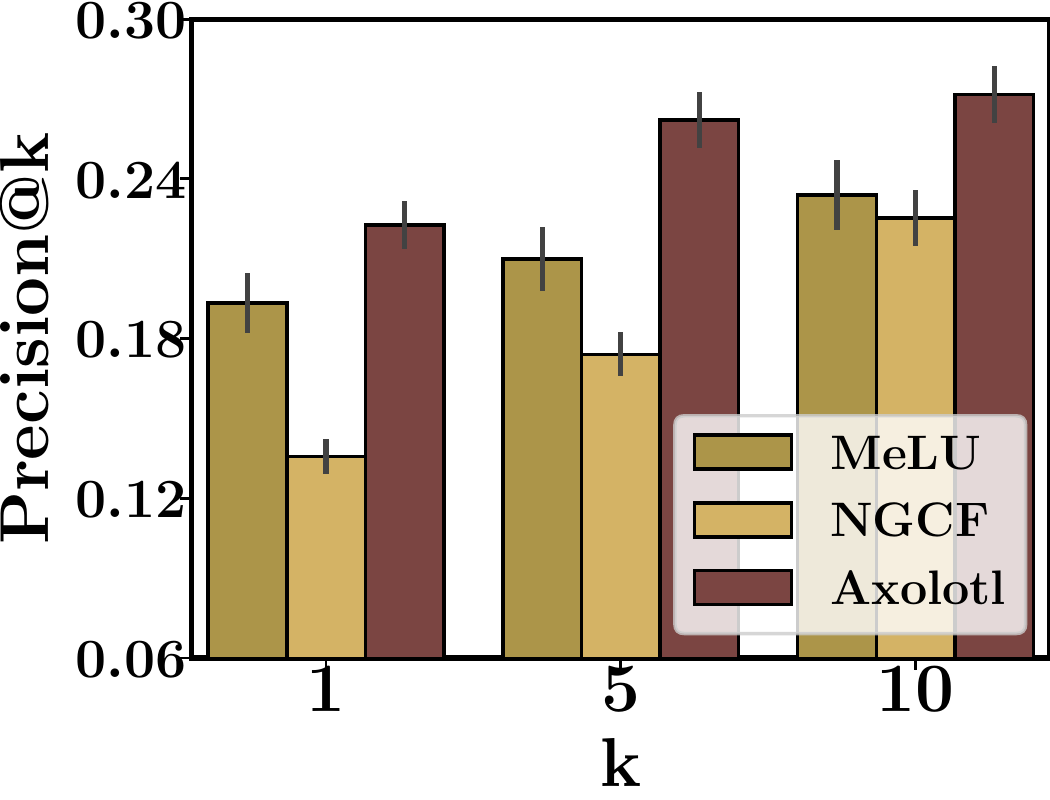}
  \caption{Washington (US)}
\end{subfigure}
\hfill
\begin{subfigure}{0.3\columnwidth}
  \centering
  \includegraphics[width=\linewidth, height=3cm]{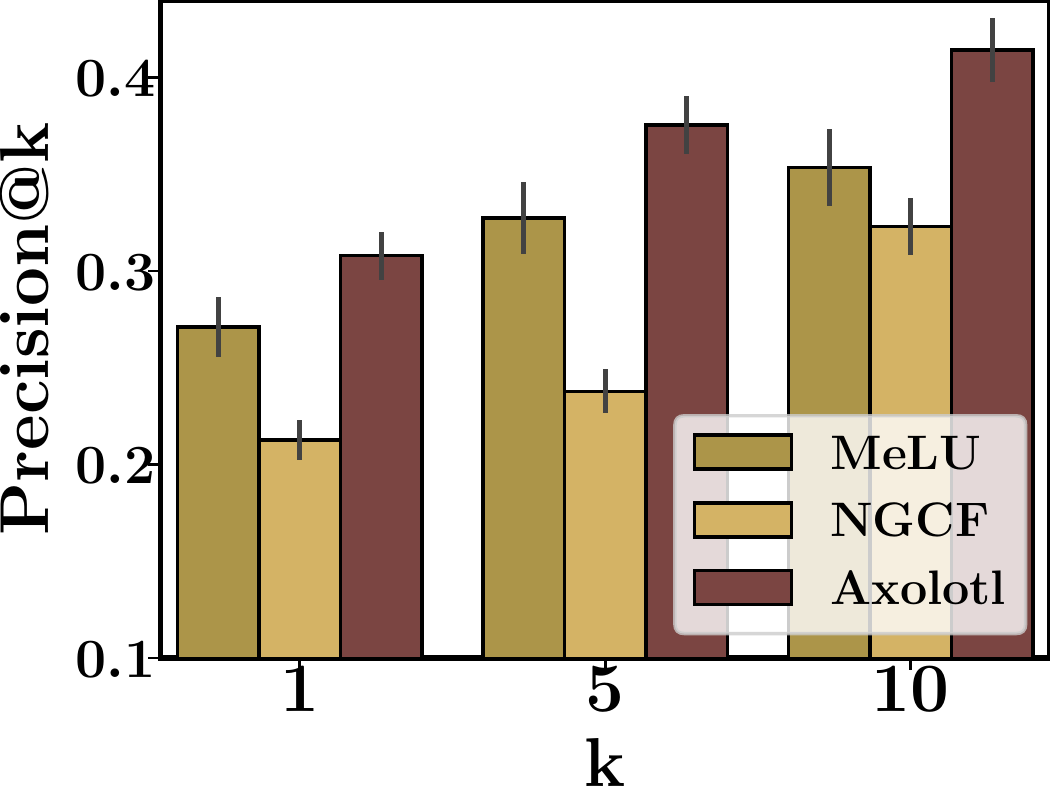}
  \caption{Aizu (JP)}
\end{subfigure}
\hfill
\begin{subfigure}{0.3\columnwidth}
  \centering
  \includegraphics[width=\linewidth, height=3cm]{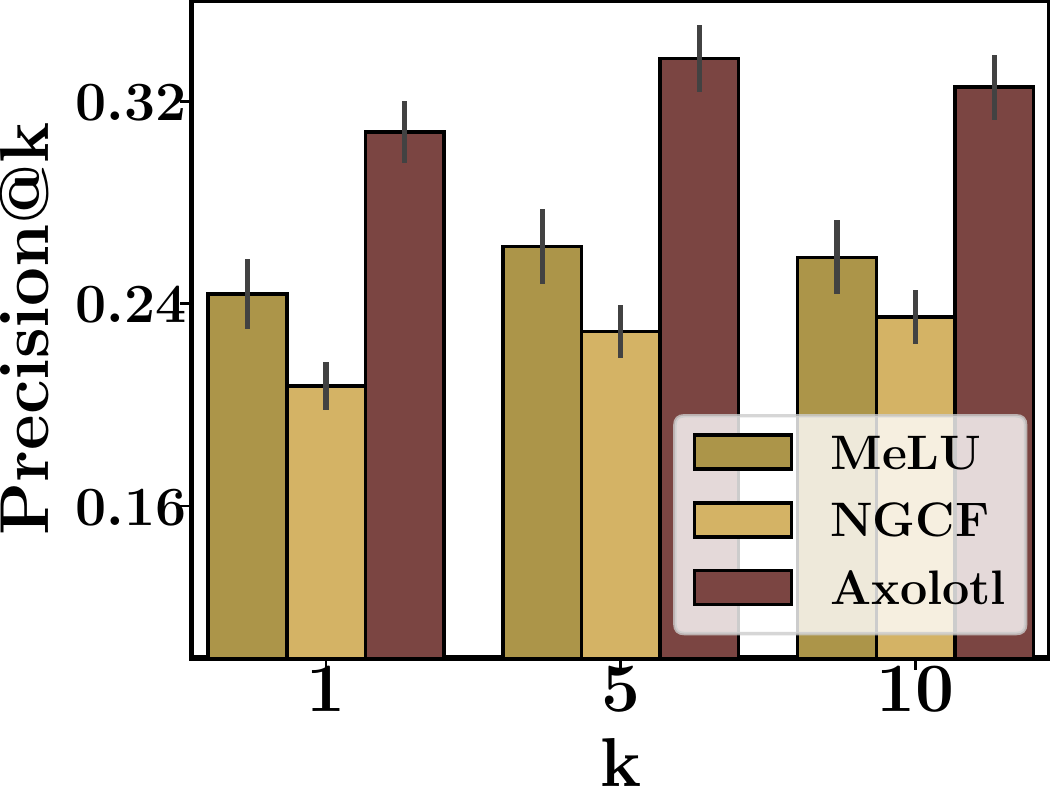}
  \caption{Berlin (DE)}
\end{subfigure}
\vspace{-3mm}
\caption{Recommending candidate POIs to a \textit{new} user in the network i.e. with no past check-in records.}
\label{fig:inductive}
\end{figure}
\subsection{Prediction Performance for New Users} 
Though the focus of this paper is recommending locations to existing users, we perform an auxiliary \textit{cold-start} recommendation experiment where we recommend candidate locations for users unseen in the training set. In this case, we obtain the final user embedding $\bs{U}_f$ of a new user based on her social network by avg-pooling~\cite{graphsage}. We compare this with the state-of-the-art recommender system, i.e. MeLU~\cite{melu} and NGCF~\cite{ngcf}. From the results in terms for precision at different thresholds in Figure \ref{fig:inductive}, we note that \ourm performs significantly better than the other approaches, thus demonstrating that \ourm is more suitable for POI recommendation even for a user with no past check-in records. We also note that MeLU easily outperforms NGCF and thus it further reinforces the need to incorporate out-of-region data for designing recommendation systems.

\begin{figure}[t]
\centering
\begin{subfigure}{0.3\columnwidth}
  \centering
  \includegraphics[width=\linewidth, height=3cm]{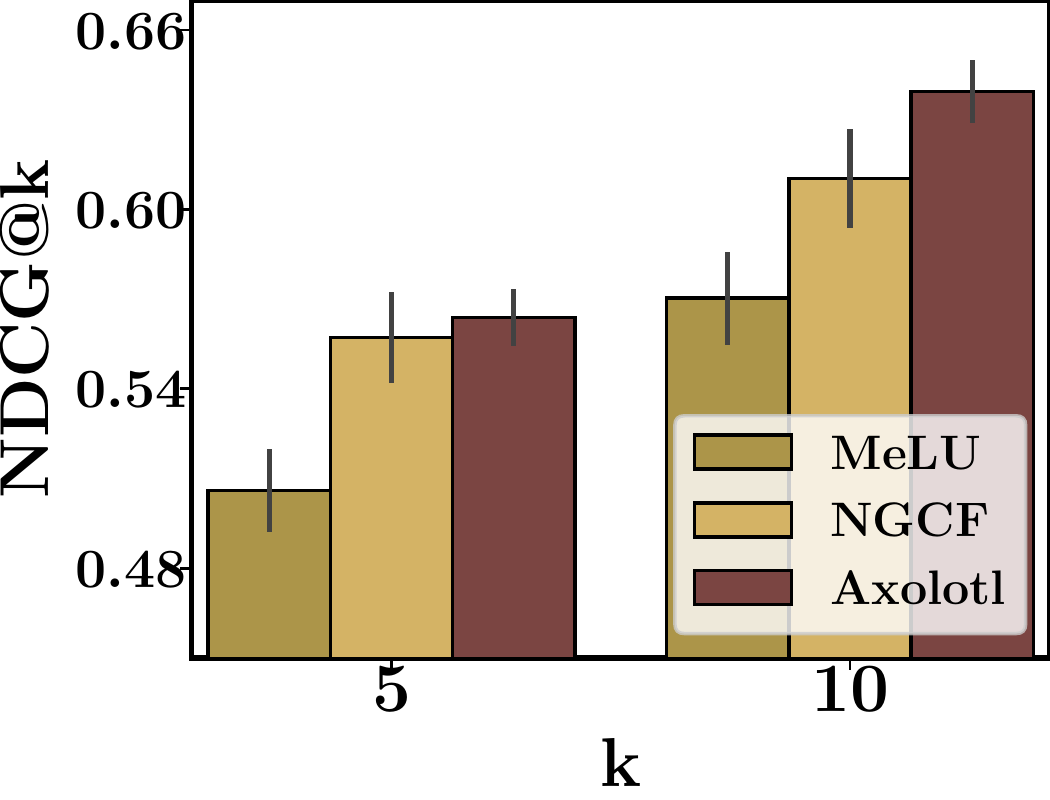}
  \caption{California (US)}
\end{subfigure}
\hfill
\begin{subfigure}{0.3\columnwidth}
  \centering
  \includegraphics[width=\linewidth, height=3cm]{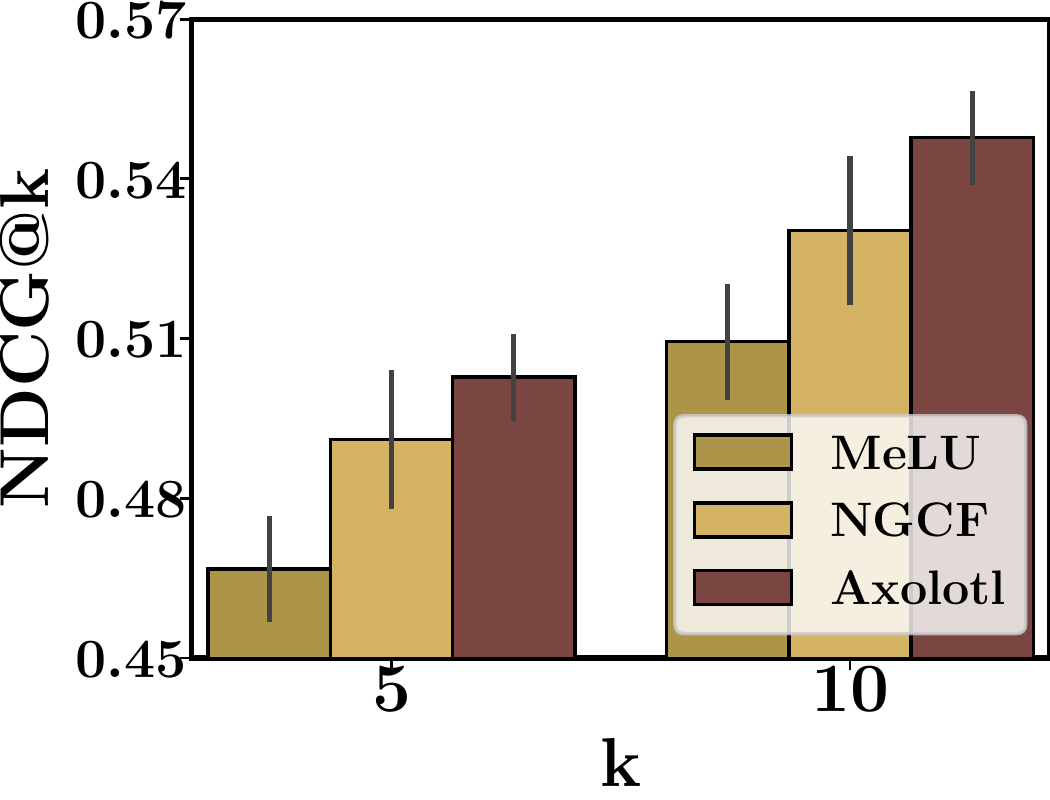}
  \caption{Tokyo (JP)}
\end{subfigure}
\hfill
\begin{subfigure}{0.3\columnwidth}
  \centering
  \includegraphics[width=\linewidth, height=3cm]{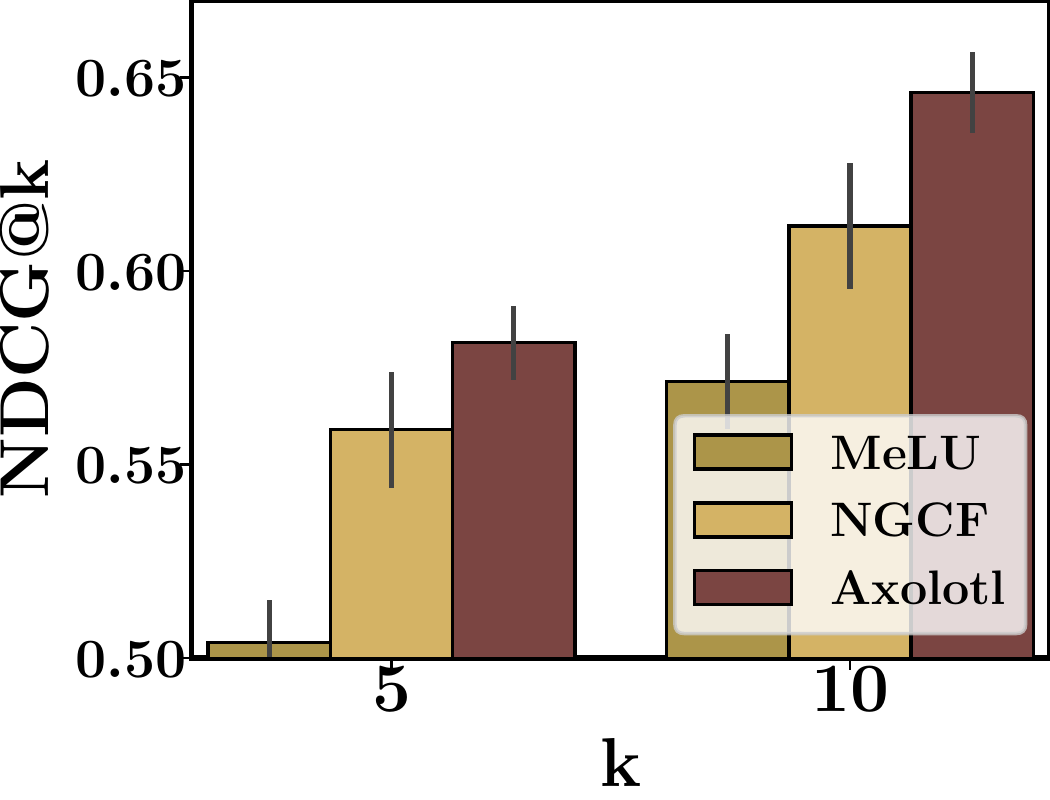}
  \caption{North-Rhine Westphalia (DE)}
\end{subfigure}
\vspace{-3mm}
\caption{Recommendation performance of Axolotl, NGCF and MeLU for users in the source region.}
\label{fig:source}
\vspace{-2mm}
\end{figure}

\subsection{\ourm for Source Regions}
We also report the recommendation performance of \ourm across different source regions, i.e. California(CA), Tokyo(TY), and North-Rhine Westphalia (NR). Here as well, we compare our model  with MeLU~\cite{melu} and NGCF~\cite{ngcf}. From the results in in Figure ~\ref{fig:source}, we note that even when recommending for users in the source region, \ourm performs comparable, if not better, than the state-of-the-art single region baseline NGCF. The results further support the practical usability of \ourm even in data-rich source regions over its ability for limited data regions. We also note that for a source region, the performance of MeLU is significantly inferior than the other two approaches.

\begin{figure}[b]
\centering
\begin{subfigure}{0.3\columnwidth}
  \centering
  \includegraphics[width=\linewidth, height=3cm]{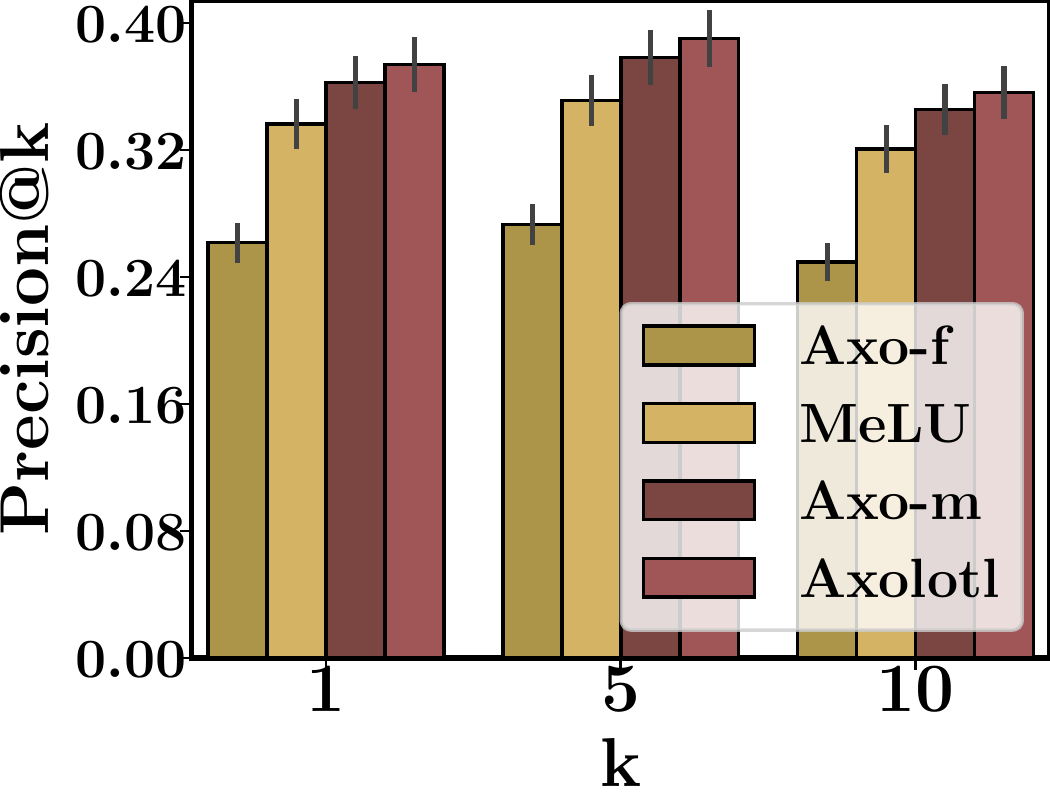}
  \caption{Washington (GOW$\rightarrow$FSQ)}
\end{subfigure}
\hfill
\begin{subfigure}{0.3\columnwidth}
  \centering
  \includegraphics[width=\linewidth, height=3cm]{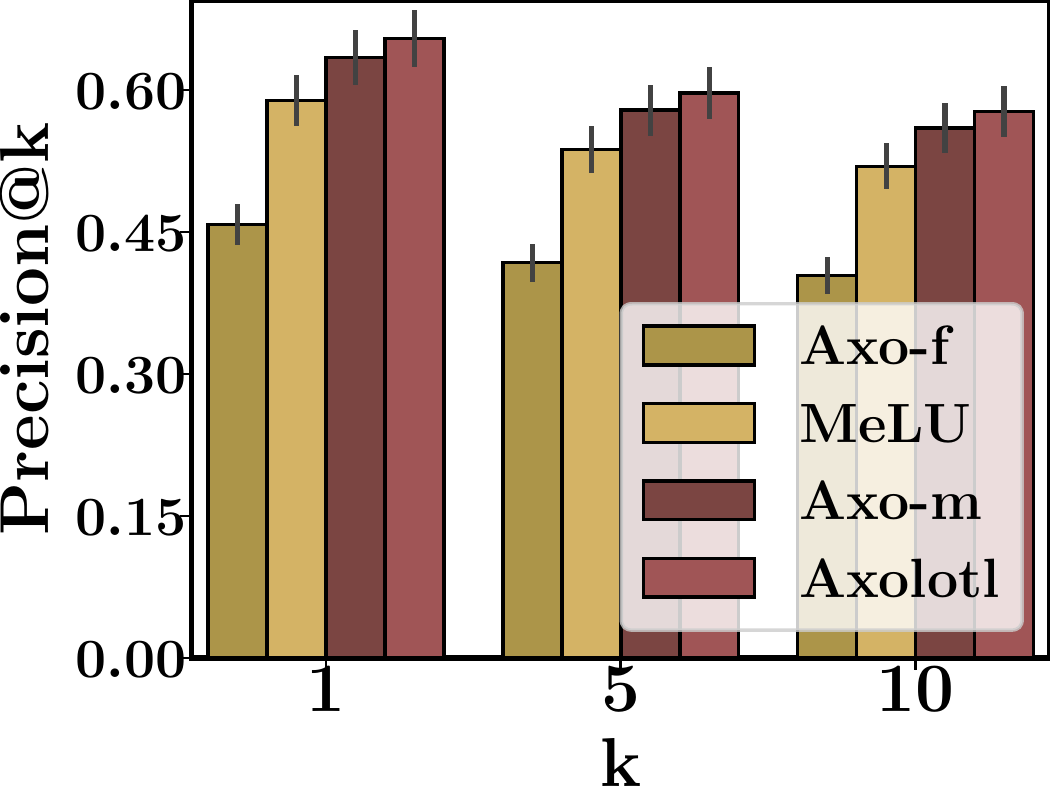}
  \caption{Aizu (FSQ$\rightarrow$GOW)}
\end{subfigure}
\hfill
\begin{subfigure}{0.3\columnwidth}
  \centering
  \includegraphics[width=\linewidth, height=3cm]{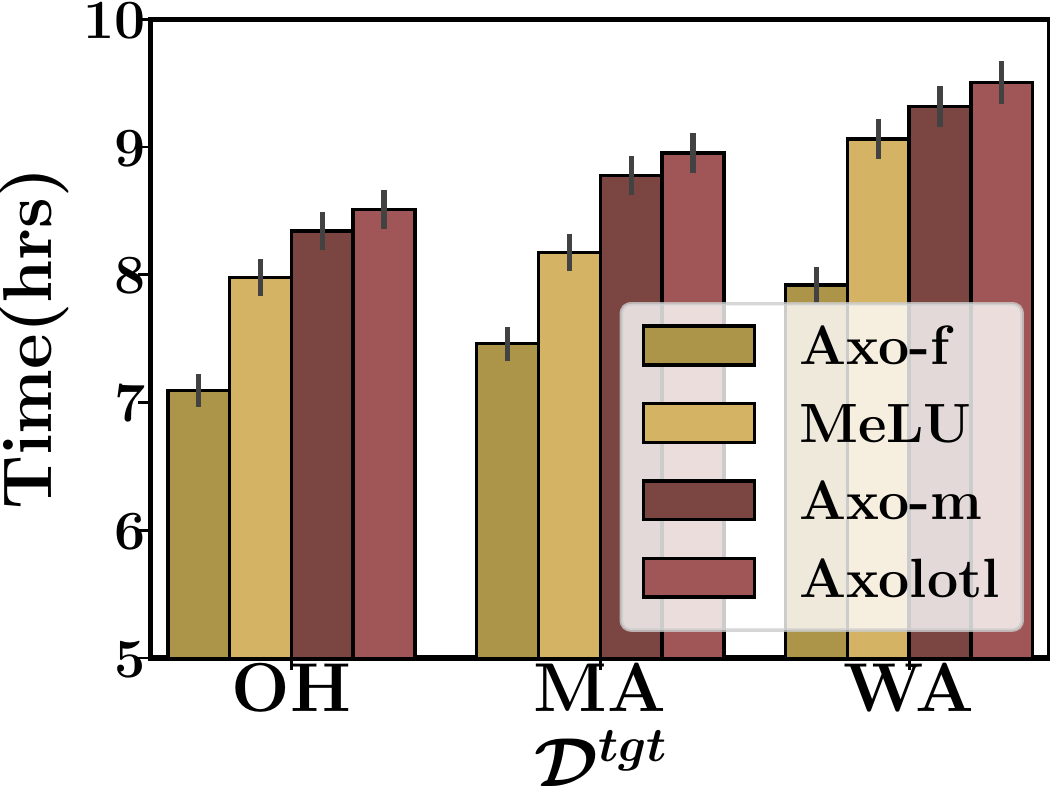}
  \caption{Training Times}
\end{subfigure}
\vspace{-3mm}
\caption{Cross-dataset recommendation performance for regions (a) Washington and (b) Aizu. (c) Training times of MeLU and different variants of \ourm for all target datasets from the US.}
\label{fig:additional}
\end{figure}

\subsection{Additional Experiments and Runtime}\label{runtime}
\noindent \textbf{Transfer across Datasets:} Finally, to further signify the ability of \ourm to even work across different datasets, we design a novel task wherein we use the check-in data from Foursquare(FSQ) and Gowalla(GOW) for a specific region, train on the data-rich variant then later test on the data-scarce variant. Specifically we perform two more experiments of (i) training our model on Gowalla (1.8k Users, 15k POIs) derived section of Washington and predicted on its Foursquare(0.6k Users, 5.2k POIs) variant, and (ii) training on Foursquare(1.9k Users, 3.4k POIs) derived variant of Aizu and test on the Gowalla(0.9k Users, 1.7k POIs) variant. In contrast to the original datasets in Table~\ref{tab:data}, here we use a more lenient criteria to filter out unnecessary users and locations, with each user having more than 2 check-ins, 2 social connections and each location with more than 3 check-ins. We plot the results for Washington and Aizu in Figure~\ref{fig:additional}(a) and Figure~\ref{fig:additional}(b) respectively. From the results we note that a standard fine tuning-based model \ours-f is not suitable for transferring across different datasets. However, meta-learning based models such as MeLU, \ours-m and \ourm perform significantly better. We also note that even when predicting across different datasets, \ourm considerably outperforms MeLU across all metrics. \\

\begin{figure}[t]
\centering
\begin{subfigure}[b]{0.3\columnwidth}
\includegraphics[width=\linewidth, height=3cm]{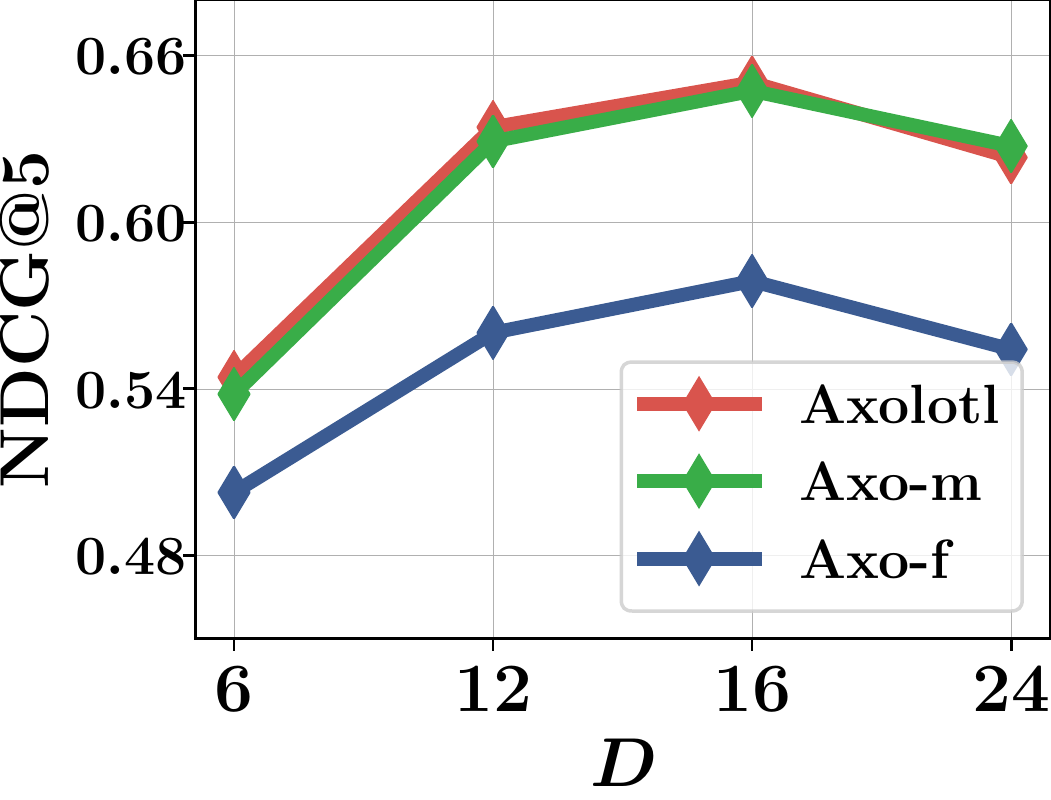}
\caption{Across Hidden Dimension}
\end{subfigure}
\hfill
\begin{subfigure}[b]{0.3\columnwidth}
\includegraphics[width=\linewidth, height=3cm]{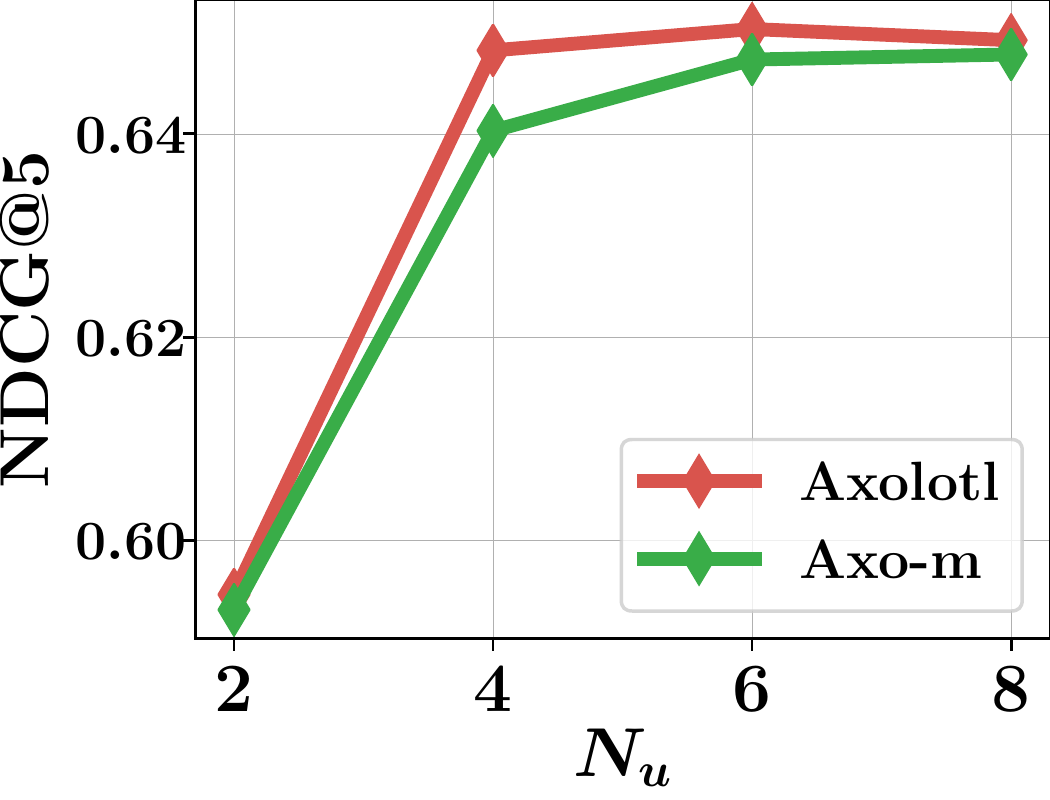}
\caption{No. of target region updates}
\end{subfigure}
\hfill
\begin{subfigure}[b]{0.3\columnwidth}
\includegraphics[width=\linewidth, height=3cm]{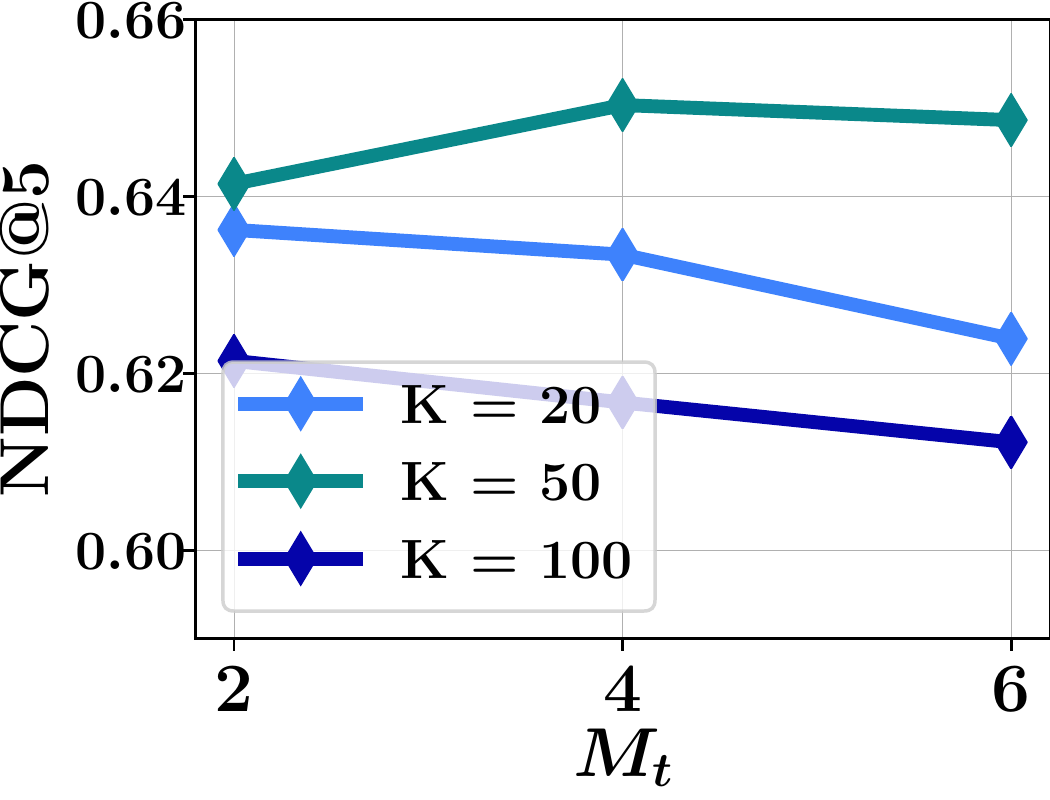}
\caption{Across $K$ and $M_t$}
\end{subfigure}
\vspace{-3mm}
\caption{\label{fig:par_wa} \ourm sensitivity across for WA dataset. Since fig(c) only considers varying cluster sizes, we omit performance of other \ourm variants and report only on the complete \ourm model.}
\vspace{-4mm}
\end{figure}
\begin{figure}[t]
\centering
\begin{subfigure}[b]{0.3\columnwidth}
\includegraphics[width=\linewidth, height=3cm]{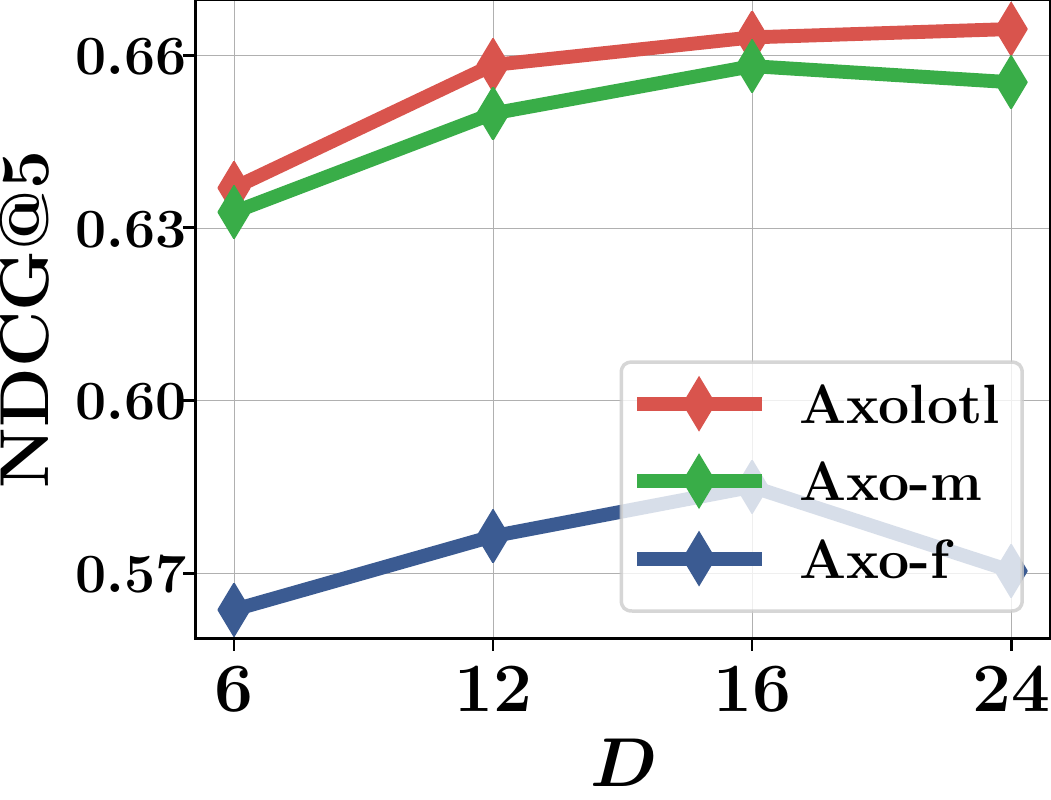}
\caption{Across Hidden Dimension}
\end{subfigure}
\hfill
\begin{subfigure}[b]{0.3\columnwidth}
\includegraphics[width=\linewidth, height=3cm]{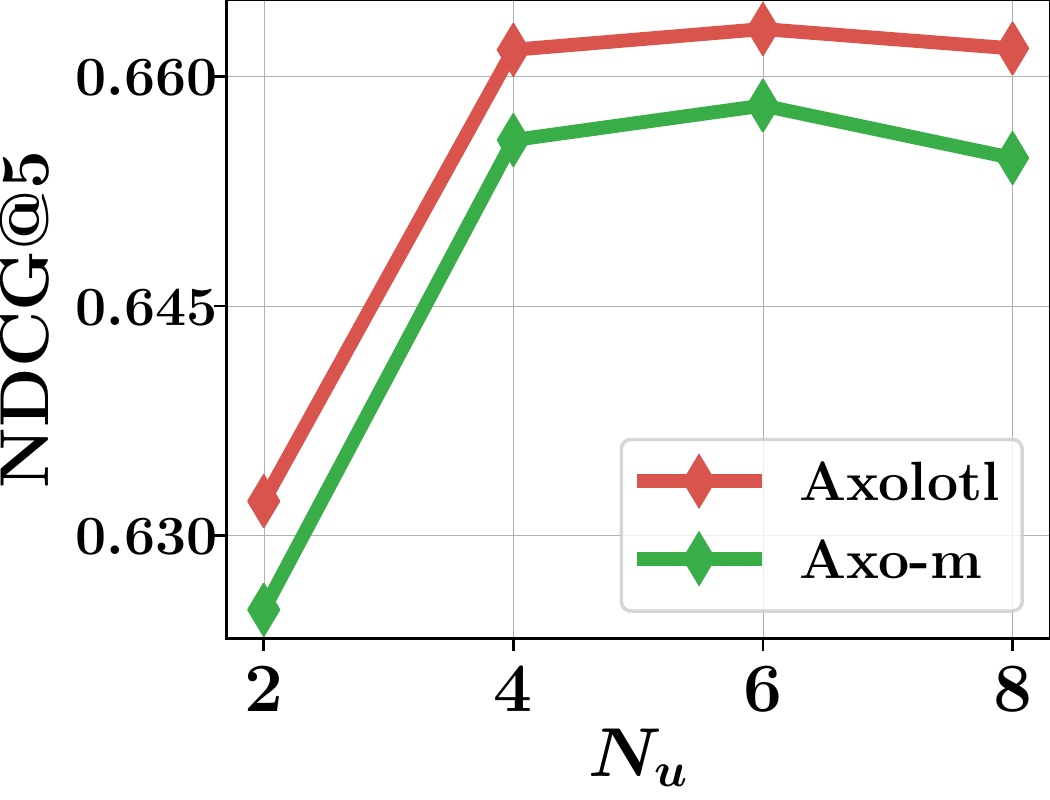}
\caption{No. of target region updates}
\end{subfigure}
\hfill
\begin{subfigure}[b]{0.3\columnwidth}
\includegraphics[width=\linewidth, height=3cm]{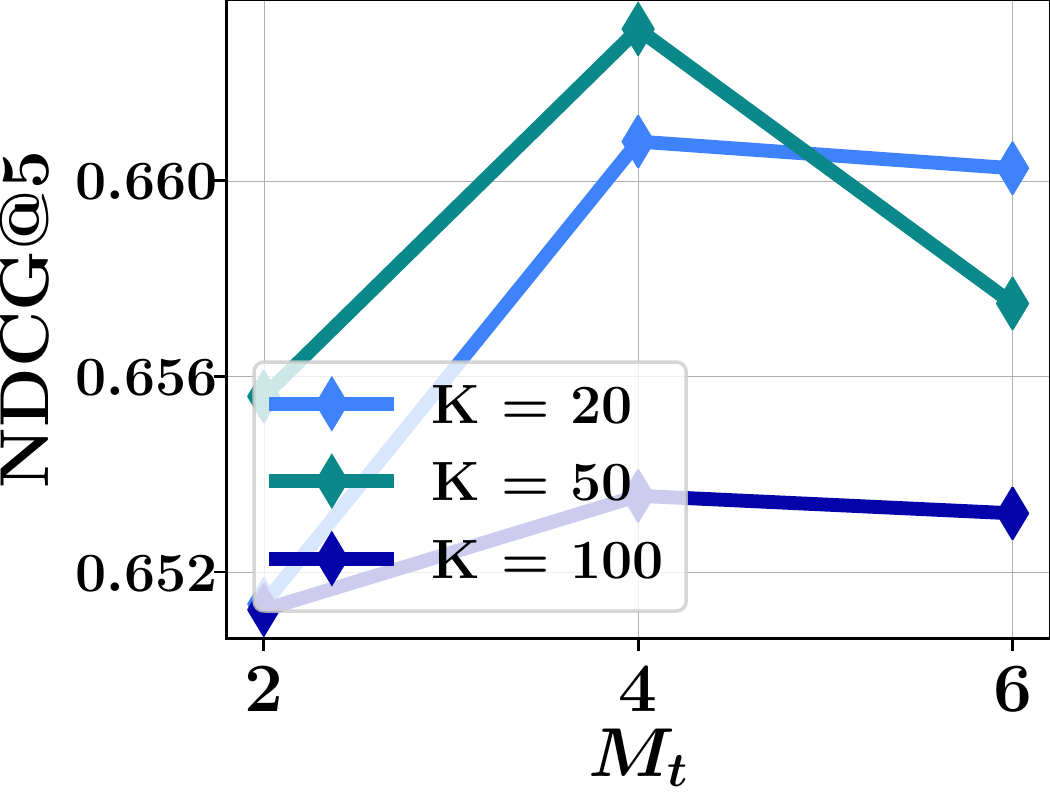}
\caption{Across $K$ and $M_t$}
\end{subfigure}
\vspace{-3mm}
\caption{\label{fig:par_ai}\ourm sensitivity across for AI dataset. Following Figure~\ref{fig:par_wa}, we only report the results of the complete \ourm model in fig (c).}
\vspace{-4mm}
\end{figure}
\begin{figure}[t]
\centering
\begin{subfigure}[b]{0.3\columnwidth}
\includegraphics[width=\linewidth, height=3cm]{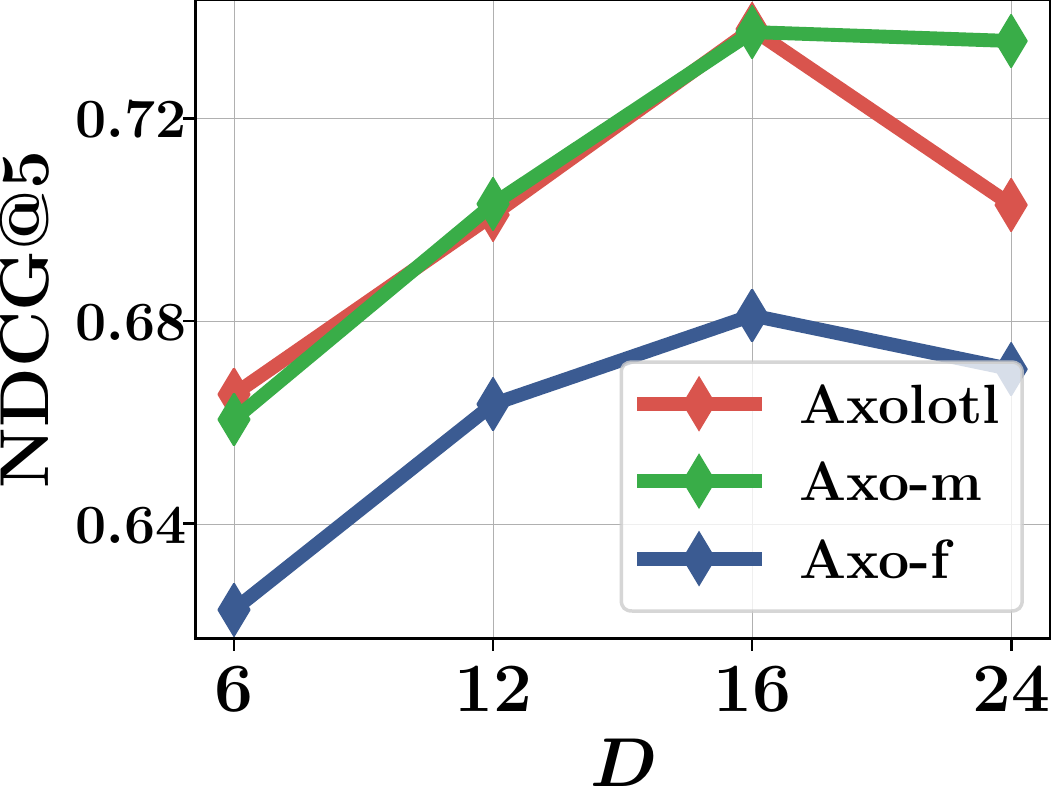}
\caption{Across Hidden Dimension}
\end{subfigure}
\hfill
\begin{subfigure}[b]{0.3\columnwidth}
\includegraphics[width=\linewidth, height=3cm]{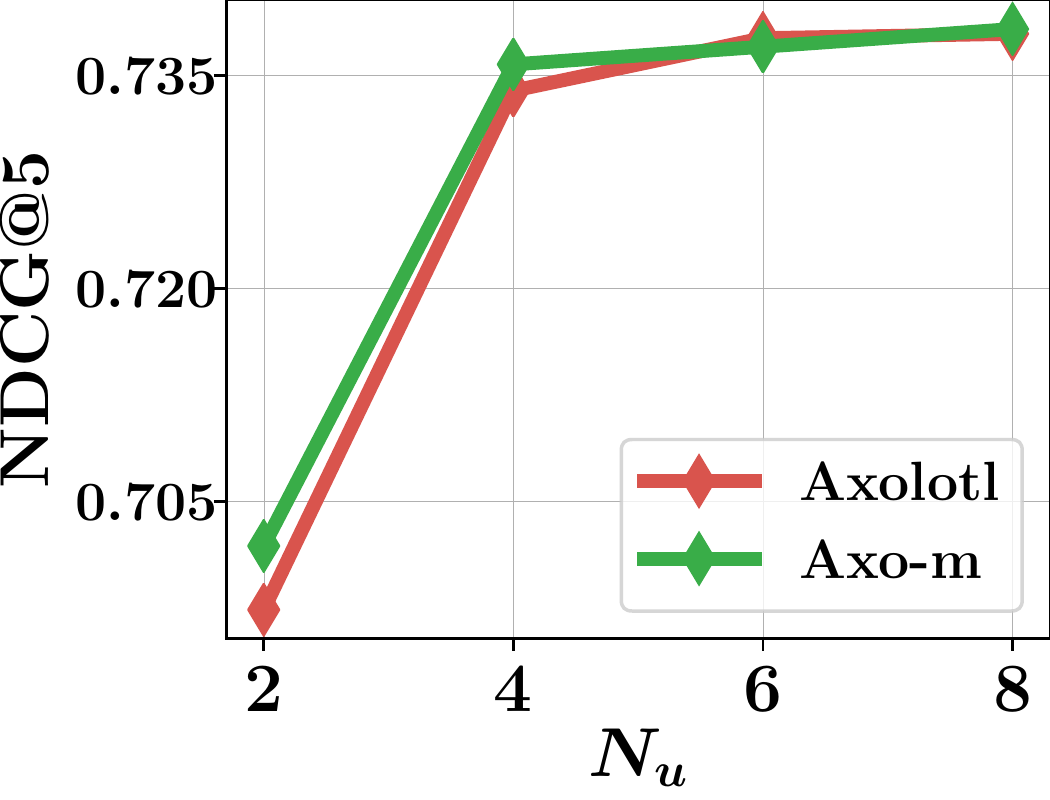}
\caption{No. of target region updates}
\end{subfigure}
\hfill
\begin{subfigure}[b]{0.3\columnwidth}
\includegraphics[width=\linewidth, height=3cm]{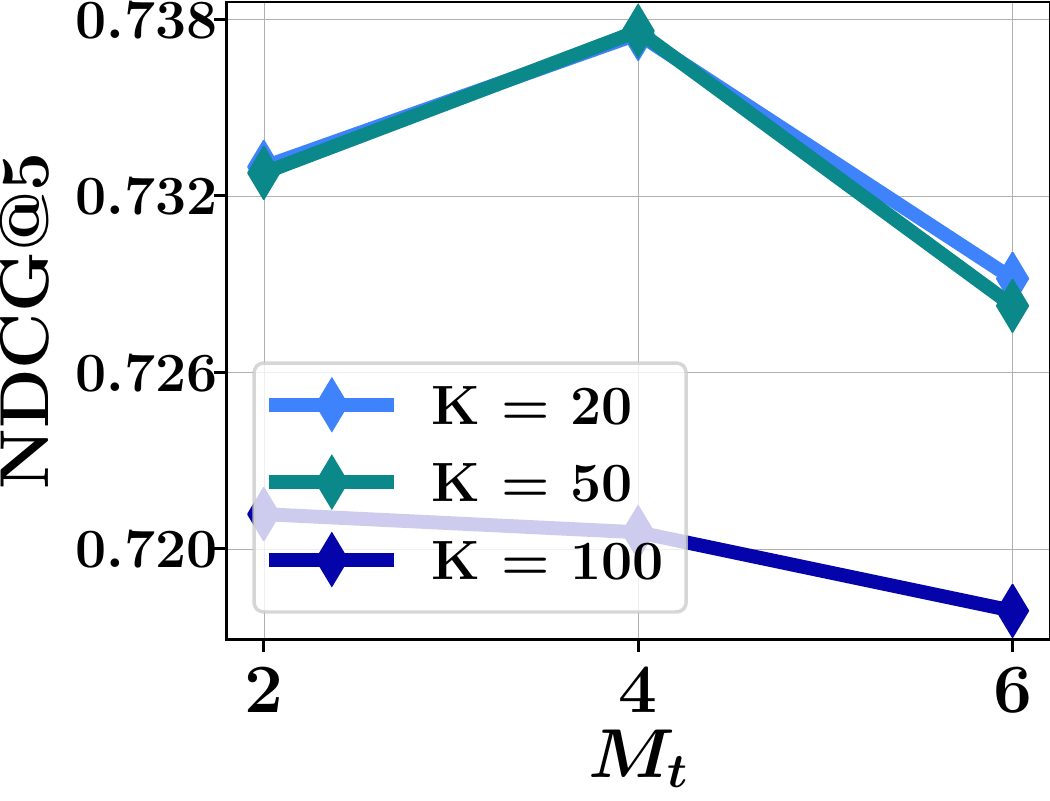}
\caption{Across $K$ and $M_t$}
\end{subfigure}
\vspace{-3mm}
\caption{\label{fig:par_be}\ourm sensitivity across for BE dataset. Following Figure~\ref{fig:par_wa}, we only report the results of the complete \ourm model in fig (c).}
\vspace{-4mm}
\end{figure}

\noindent \textbf{Runtime Analysis:} We also report on the run-time of \ourm to verify its applicability in real-world settings. We report the run-times for \ourm over the largest datasets, i.e. the states in the U.S. CA(42.1k POIs) $\rightarrow$ WA(16.7k POIs), MA(10.5k POIs), OH(8.5k POIs). From the results in Figure ~\ref{fig:additional}(c), we note that even for transferring across largest dataset, i.e Washington, the overall training time of \ourm on a Tesla V100 GPU is comparable to those of MeLU. We also highlight that these differences were even lower in smaller datasets which we exclude for brevity. Though the training times are feasible for deployment, the large run-time is mainly due to the inefficient CPU based batch-sampling. With a GPU-based sampling alternatives~\cite{fastgcn, gcngpu, dac}, this run-time can be significantly improved which we consider as a future work. Though the current single-threaded sampling affects the training time of \ourm, the time-taken (in seconds) for \textit{inference}, being: 3.27, 3.19, and 5.74 for Aizu, Berlin and Washington respectively, are well within range for practical deployment. We also note that the cluster-alignment loss improves the prediction performance of \ourm and yet includes a minor computation cost over the meta-learning variant \ours-m.

\subsection{Parameter Sensitivity (RQ4)} \label{sensitive}
Finally, we perform the sensitivity analysis of \ourm over key parameters: (1) $D$, the dimension of embeddings; (2) $N_u$, no. of meta-updates on a target batch before a source batch update; (3) $M_t$, epochs after which we initiate cluster based transfer; and (4) $K$, no. of clusters for source and target (see Table \ref{tab:par}). We report the results for WA, AI and BE in Figure~\ref{fig:par_wa},~\ref{fig:par_ai}, and~\ref{fig:par_be} respectively. We note that as we increase the embedding dimension the performance first increases since it leads to better modeling. However, beyond a point, the complexity of the model increases requiring more training to achieve good results, and hence we see its performance deteriorating. Across $N_u$, we note that increasing the number of updates with batches from the target region leads to better results before saturating at a certain point. We found $N_u = 4$ to be the optimal point across datasets in our experiments. Across $M_t$, we notice that a delayed transfer based on clusters leads to a degradation in performance. This could be attributed to the saturation of parameters due to continuous updates through meta-learning iterations. Finally across $K$, we observe the presence of an optimal number of clusters for the source and target regions. Specifically, increasing $K$ leads to a better recommendation performance as it provides higher flexibility to POI and users with similar characteristics to be assigned the same cluster while simultaneously discarding the noisy entities. However, we note that beyond a certain threshold the performance deteriorates as it reduces the variance between clusters and creates artificial boundaries within similar clusters.

\section{Conclusion}
\label{sec:concl}
In conclusion, we developed a novel architecture called \ourm that incorporates mobility data from other regions to design a location recommendation system for data-scarce regions. We also propose a novel procedure called spatio-social meta learning approach that captures the regional mobility patterns as well as the graph structure. We address the problems associated with an extremely data-scarce region and devise a suitable cluster-based alignment loss that enforce similar embeddings for communities of user and locations with similar dynamics. Our novel graph attention based model captures the complex user-location influence patterns for a specific region. Experiments over diverse mobility datasets revealed that \ourm is able to significantly improve over the state-of-the-art baselines for POI recommendation in limited data regions and even performs considerably better across datasets and data-rich source regions. As a future work for this paper, we plan to modify the transfer procedure of \ourm to incorporate novel meta-learning approaches, namely ProtoMAML~\cite{protomaml} and Reptile~\cite{reptile}. Additionally, we plan to experiment with approaches~\cite{sagpool, diffpool, spat} that automatically identify the number of clusters within a region while simultaneously maintaining the scalability of our model.

\begin{acks}
This work was partially supported by an IBM AI Horizons Network (AIHN) grant and a DS Chair of AI fellowship to Srikanta Bedathur. 
\end{acks}

\bibliographystyle{ACM-Reference-Format}
\bibliography{main}
\end{document}